\documentclass{article}
\usepackage{jfrExamplee}
\usepackage{graphicx}
\usepackage{apalike}
\usepackage{setspace}
\usepackage{amsmath} % assumes amsmath package installed
\usepackage{amssymb}  % assumes amsmath package installed
\usepackage{float}

\newcommand{\boldalpha}{\boldsymbol\alpha}
\newcommand{\boldkappa}{\boldsymbol\kappa}
\newcommand{\boldlambda}{\boldsymbol\lambda}
\newcommand{\boldmu}{\boldsymbol\mu}
\newcommand{\boldomega}{\boldsymbol\omega}
\newcommand{\psidown}{\psi_\text{down}}
\newcommand{\psiup}{\psi_\text{up}}
\newcommand{\psiditch}{\psi_\text{ditch}}
\newcommand{\psimound}{\psi_\text{mound}}
\newcommand{\degreeb}{\ensuremath {}^{\circ}}

\title{Autonomous Multirobot Excavation \\for Lunar Applications}

\author{
Jekanthan Thangavelautham \\
School of Earth and Space Exploration\\
Arizona State University\\
781 E Terrace Mall, Tempe, AZ 85287 \\
\texttt{jekan@asu.edu} \\
\And
Kenneth Law \\
David Schaeffer and Associates \\
Markham, ON, Canada \\
\And
Terence Fu\\
University of Toronto\\
4925 Dufferin Street, Toronto, Ontario, Canada M3H 5T6 \\
\And
Nader Abu El Samid \\
MDA Space Missions \\
9445 Airport Road \\
Brampton, ON, Canada L6S 4J3 \\
\And
Alexander D.S. Smith \\
University of Toronto \\
4925 Dufferin Street, Toronto, Ontario, Canada M3H 5T6 \\
\texttt{alexander.smith@utoronto.ca} \\
\AND
Gabriele M.T. D'Eleuterio \\
University of Toronto \\
4925 Dufferin Street, Toronto, Ontario, Canada M3H 5T6 \\
\texttt{gabriele.deleuterio@utoronto.ca} \\
}

\begin{document}

\maketitle

\begin{abstract}
In this paper, a control approach called Artificial Neural Tissue (ANT) is applied to multirobot excavation for lunar base preparation tasks including clearing landing pads and burying of habitat modules.  We show for the first time, a team of autonomous robots excavating a terrain to match a given 3D blueprint.  Constructing mounds around landing pads will provide physical shielding from debris during launch/landing. Burying a human habitat modules under 0.5 m of lunar regolith is expected to provide both radiation shielding and maintain temperatures of -25 $^{o}$C.  This minimizes base life-support  complexity and reduces launch mass.  ANT is compelling for a lunar mission because it doesn't require a team of astronauts for excavation and it requires minimal supervision.  The robot teams are shown to autonomously interpret blueprints, excavate and prepare sites for a lunar base.  Because little pre-programmed knowledge is provided, the controllers discover creative techniques.  ANT evolves techniques such as slot-dozing that would otherwise require excavation experts.  This is critical in making an excavation mission feasible when it is prohibitively expensive to send astronauts.
The controllers evolve elaborate negotiation behaviors to work in close quarters. These and other techniques such as concurrent evolution of the controller and team size are shown to tackle problem of antagonism, when too many robots interfere reducing the overall efficiency or worse, resulting in gridlock.   While many challenges remain with this technology our work shows a compelling pathway for field testing this approach.
%\keywords{collective robotics \and excavation \and neural networks \and evolutionary algorithms }
\end{abstract}

\section{Introduction}

Social insects such as a colony of ants excavating a network of tunnels or swarms of termites building towering cathedral mounds with internal heating and cooling shafts \cite{bristow} show the potential of multi-agent systems in building robust structures.   These social insects, without any centralized coordination produce emergent collective behaviors used to build these robust structures.  Multiple individuals working in a decentralized manner offers some inherent advantages, including fault tolerance, parallelism, reliability, scalability and simplicity in robot design \cite{cao}.

\begin{figure} [h]
\centering
\includegraphics[width=4in]{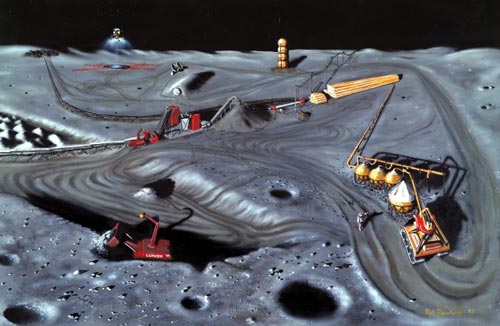}
% \vspace{-12pt}
\caption{Artist Impression of a Lunar Base and Mining Facility (courtesy NASA).}
% \vspace{-8pt}
 \label{fig:lunar_mine}
\end{figure}

Using this bio-inspired method, teams of autonomous robots can construct key elements of a human habitat on the moon (Figure~\ref{fig:lunar_mine}).  They can work continuously in harsh environments making them very productive, are fault tolerant to the failure of individual robots and are scalable depending on the task complexity and schedule. Robots do not require life-support infrastructure that would otherwise be required for a team of astronaut workers.  Furthermore, robots may be required for certain tasks due to concerns of health and safety of the astronauts. Combining these factors, our studies show use of teams of autonomous robots instead of astronauts  can reduce launch cost by 50\% \cite{nader2}  for lunar base construction.  This can free ground operators from constantly tending to the multirobot team. In this architecture it will be possible for operators on the ground to have total oversight over the activities of the robot team and intervene and recover from mishaps or unexpected events.

Constructing mounds around landing pads will provide physical shielding from debris during launch/landing. Burying a human habitat modules that under 0.5 m of lunar regolith is expected to provide both radiation shielding and maintain comfortable temperatures of -25 $^{o}$C (based Apollo 17 manual excavation experiments)\cite{heiken}.

Earth-based teleoperation systems have been proposed for control of robots on the moon \cite{spong}.  Such systems have been demonstrated successfully with the Lunakhod 1 and 2 rover missions; however latency (time delay) induced prolonged fatigue was a concern for the Lunakhod missions.  Latency induced operator fatigue is still a concern   when coordinating actions with teams of robots over extended time \cite{miller}.  Advancements have been made coordination and control of multiple robots using teleoperation \cite{lu,khad}. However these techniques have yet to be tested for spacecraft or robots at the Moon.

The proposed ANT architecture allows ground control oversight and frees ground operators from mundane tasks while reserving all but the most delicate and mission critical tasks for human intervention thus reducing the chance of fatigue and human error.  These factors make an autonomous robotic system with teleoperation capability more appealing than teleoperation alone. This approach permits having a base deployed and operational in time for astronauts to arrive from Earth. Major approaches to developing autonomous control systems utilize human knowledge, machine learning techniques or a combination of both.

Human knowledge-based control strategies rely on human input in the form of ad-hoc control rules, physical model based planners, task-specific assumptions, and human knowledge \cite{bernd,xid}.  In many tasks that use multiple robots, it is often unclear of how best to plan the task,  organize the group and determine the best interaction behaviors to complete the task.  In lunar and planetary environments, task-specific assumptions may not be valid in-situ.  The surface properties and material may vary from crater to crater.  One or more robots may be disabled unexpectedly or need to perform tasks that have not been envisioned during mission planning and modeling stages.  These factors make an adaptive, decentralized controls approach to reorganize and control a group of robots more appealing.

A novel, robotic learning controls approach is presented here that addresses these challenges.  This approach requires much less human knowledge than conventional approaches. The controllers are homogenous (i.e., a single controller is replicated for use on each robot), decentralized and make use of local sensor data.  Hardware implementations of the system utilize a shared resource such as an overhead camera for localization or TriDAR for 3D mapping and hence the system is not truly decentralized but the system can be made decentralized by utilizing multiple shared resource. The approach learns to solve tasks through a process of trial and error and is given a global objective function, a generic set of basis behaviors, sensory input and a simplified training environment without detailed physical models of the system.  The proposed approach requires an accurate localization system.  This is possible by mounting cameras and lighting on a tower over the work area.  Other options include use of radio beacons that are located from the main landing craft or structure.  These two approaches can enable rover localization without requiring a Lunar GPS system.

This approach called the Artificial Neural Tissue (ANT)  \cite{Thangav2005,Thangav2008} combines a standard neural network with a novel coarse-coding mechanism
inspired by the work of Albus and Hinton \cite{albus,hinton}. In ANT, coarse coding is used to perform regulatory control of networks \cite{Thangav2005,Thangav2008,Thangav2010}. The process occurs through simulated chemical diffusion \cite{garth,montague} and enables segmentation of a network, through activation and inhibition of neuronal groups.  This allows for modules of neurons to be added, removed and rewired during training facilitating self-organized task decomposition and  task-allocation \cite{Thangav2010}. This method is shown to solve the sign-following task found to be intractable with conventional fixed and variable topology neural network architectures \cite{Thangav2005,Thangav2010}.

Here the capabilities of ANT are shown for multirobot excavation \cite{thangavstaif,ThangavCIRA2009}, a difficult task, with a large, high-dimensional task space. Learning to solve the excavation task, enables the robots to build berms, landing pads and excavate holes for burying the lunar habitat modules.  The excavation task combines features of a typical foraging, grazing or cleaning task with ability to plan, interpret blueprints and perform coordinated excavation. Since little pre-programmed knowledge is given, ANT may discover creative solutions producing controllers that can interpret excavation blueprints, can successfully avoid obstacles, perform layered digging, leveling and avoid burying or trapping other robots. These innovative behaviors are discovered during training and are similar to behaviors used by social insects to perform group coordination.

In this work expanded from  \cite{thangavstaif},  ANT is found to evolve superior solutions to the excavation task in fewer genetic evaluations than hand-coded and conventional neural networks  solutions.  The ANT solutions are found to be scalable and can be applied to unforseen real world scenarios where one or more robots may become disabled or unavailable.  Hand-coded solutions are found to work in single robot scenarios and show poor performance, and robustness for increased number of robots thus lacking adaptivity.  The required cooperative behaviors for all but the simplest of tasks are unintuitive and pose difficulty for humans programmers. Conventional neural networks can do better than hand coded solutions but require an experimenter manually decompose a complex task,  determine a suitable network topology and activation function to make training tractable.  ANT requires even less experimenter input and is useful for multirobot excavation tasks, where there is limited domain knowledge available. The controllers can generalize (interpolate) from limited training scenarios to handle unforseen situations.   ANT through a process of coarse-coding can segment high dimensional tasks space more efficiently, performing automated task decomposition and simultaneously finding the required controller network topology, selecting optimal number of robots and coordination behaviors to complete the task.  Neuronal activity and behavioral analysis of the controllers suggests solutions emerge through a process of trial and error discovery, fine tuning and incremental assembly of `building-block' behaviors to solve tasks.

Using this approach we show the feasibility of using multirobot excavation
for site preparations tasks.  The approach shows improved performance and scalability than conventional neural networks and hand coded solutions.   This facilitates finding creative behaviors that are not specified or encouraged by an experimenter.  These creative behaviors verified in hardware include correctly interpreting blueprints, performing layered digging, obstacle avoidance and rocking behaviors to avoid getting stuck.  This approach is shown to produce controllers that have improved scalability compared to conventional neural networks and hand-coded solutions.   Furthermore, ANT can simultaneously evolve the desired controller and select for optimal number of robots for the task. This approach is shown as a possible solution to the problem of antagonism in decentralized multirobot control.

The evolved solutions have been analyzed in simulation and the best solutions have been ported onto real robots.  Hardware experiments were performed on 3 different robotic platforms, including in the laboratory and under controlled field conditions.  These experiments were used to validate individuals behaviors seen in simulation to verifying the overall excavation performance of the controllers. Laboratory hardware experiments and controlled field experiments produced promising results that show a promising pathway towards full implementation and demonstration in the field.

This article is organized as follows.  Section 2 presents related work. Section 3 presents the Artificial Neural Tissue approach. Section 4 presents the excavation task used to demonstrate ANT's capabilities.  This is followed by results and discussion in Section 5 and proof-of-concept experiments in Section 6.

\section{Related Work}
\label{ch3:related_work}

Previous work in autonomous excavation \cite{stentz} has been limited to a single robot and separate loading/unloading vehicles. Digging is performed using hand coded scripts that simplify repetitive excavation/truck loading cycles.
These controllers are used to position and unload an excavator bucket relative to a dump truck using a suite of sensors onboard the vehicles.

These scripts are developed with input from an expert human operator and model vehicle specific
limitations such as load handling capacity and latency. These systems incorporate adaptive  coarse and refined planners to sequence
digging operations  \cite{Rowe1997}.  Other works used coarse and refined planner containing a neural network approach to learn soil conditions and excavator capabilities during operation \cite{Cannon1999}. Such systems are comparable in efficiency to human operators.  Other approaches are devoted to modeling kinematics and/or dynamics of the excavation vehicles  and simulating their performance \cite{dunbabin}. These techniques are designed for specific vehicle platforms and do not include scripts for every possible scenario in the task space thus requiring close monitoring and intervention by a human operator. This makes the approach unsuitable for fully autonomous operation of multiple robots on the moon.

Control systems such as LUCIE  are more sophisticated and incorporate long-term planning \cite{bradley}.
Apart from identifying and automating excavation cycles, the system incorporates a whole sequence of intermediate goals that need to be achieved to complete a trench digging task. However, the system lacks autonomy because the task is decomposed and prioritized by a human operator.

Other techniques closely mimic insect in there ability to build road ways and ramps using amorphous construction.  A laboratory robot is used to heat, melt and deposit foam to produce a ramp and other complex structures \cite{napp}.  More recent work by Halbach et al. \cite{halbach} have performed simulations of multiple robots to perform excavation on Mars.  The intent is to setup a permanent human base and utilize Martian resources for construction and in-situ resource utilization.  The work has focused on human assisted high level planning required to  locate a base and key facilities and the process of resource extraction.  Human assistance is utilized in planning the high level tasks and giving execution orders to the multiple robots.  It is presumed human astronauts are already located on Mars and can perform tele-operation on site (from a safe distance).  Recent work by \cite{skon} have show bucket wheels to be the most effective excavation platform for low gravity on the Moon.  As will be shown later, our results also show bucket wheels to be most efficient for excavation.

The construction of a human habitat on the moon will require multiple excavation robots working towards a given goal.  Collective robotics is well suited because it incorporates multiple autonomous robots that work cooperatively towards a global goal. Some collective robotic controllers mimic mechanisms used by social insects to perform group coordination.  These include the use of self-organization, templates and stigmergy. \emph{Self-organization} describes how macroscopic behavior emerge solely from numerous interactions among lower level components of the system that use only local information~\cite{Bonabeau1997} and is the basis for the bio-inspired control approach presented here.  Templates are environmental features perceptible to individuals within the collective~\cite{Bonabeau1999}.   In robotic applications, template-based approaches include use of light fields to direct the creation of linear \cite{Stewart2003} and circular walls \cite{Wawerla2002} and planar annulus structures \cite{Wilson2004}. Spatiotemporally varying templates (\emph{e.g.}, adjusting or moving a light gradient over time) have been used to produce more complex structures \cite{Stewart2004}.

Stigmergy is a form of indirect communication mediated through the environment~\cite{Grasse1959}. Stigmergy has been used extensively in collective-robotic construction, including blind bulldozing \cite{Parker2003}, box pushing \cite{Mataric1995},  heap formation \cite{Beckers1994} and tiling pattern formation \cite{Thangav2003}.

However conventional collective robotics control approaches have two limitations.  First, they rely on either user-defined
deterministic ``if-then'' rules or on stochastic behaviors.    It is difficult to design controllers by hand with cooperation in mind, as we show later in the paper, because there exists no formalisms to predict or control the global behaviors that will result from local interactions. Designing successful controllers by hand can devolve into a process of trial and error.

The second limitation is that these approaches can suffer from an emergent feature called \emph{antagonism}~\cite{chante} when multiple agents trying to
perform the same task interfere with one another, reducing the overall efficiency of the group or worse, result in gridlock.  This limits scalability of the solution to number of robots and size of the task area.

Because the approach presented here is evolutionary in nature, it ``learns'' to exploit the mechanisms described earlier to find creative solutions to a task.  As with other evolutionary algorithms, the approach is stochastic  and cannot guarantee a solution in finite time.  However, as will be presented later, the controllers converge to solution with a probability of 93\% at the optimal training settings. The presented method is able to mitigate the effects of antagonism, which is difficult to do with conventional approaches due to lack of domain knowledge of a task at hand.

A means of reducing the effort required in designing
controllers by hand is to encode controllers as behavioral look-up
tables and allow a genetic algorithm to evolve the table entries.
This approach is used to solve a heap formation task in
\cite{Barfoot1999} and a $2 \times 2$ tiling formation task in
\cite{Thangav2003}.

A limitation with look-up tables is that they have poor sensor
scalability, as the size of the look-up table is exponential in the
number of inputs. Look-up tables also have poor generalization.
Neural network controllers perform better generalization since they
effectively encode a compressed representation of the table. Neural networks have been successfully applied on multirobot systems and have been used to build walls, corridors, and briar patches \cite{Crabbe1999} and for tasks that require communication
and coordination \cite{Trianni2006}.

Neural network
controllers have been also been used to solve $3 \times 3$ tiling task
\cite{Thangav2004}.  Going from the $2 \times 2$ to the $3 \times 3$ tiling formation task, results in a search space of $10^{145}$ to $10^{1300}$ respectively.  This very large increase in search space prevents a lookup table from finding a suitable solution.  However because a neural network can generalize better it finds a desired solution.

In standard neural networks, communication between neurons is modeled as synaptic connection (wires) that enable electrical signalling.  Other fixed topology networks such as Gasnet model both electrical and chemical signalling between neurons \cite{husband}. However, when using fixed-topology networks, the size of the network must be specified ahead of time.
Choosing the wrong topology may lead to a network that is difficult to train or is intractable \cite{jordan:jacob,Thangav2005}.

Variable length neural network methodologies such as NEAT (NeuroEvolution of Augmenting Topologies) show the potential advantage of evolving both the neuron weights and topology concurrently \cite{kstanley}. It is argued that growing the network incrementally through  (`complexification') helps minimize the dimensional search space and thus improve evolutionary performance \cite{kstanley}.  This requires starting with a small topology and growing it incrementally through evolutionary training which can be slow.

The ANT framework presented here is a bio-inspired approach that simultaneously addresses both the problems in designing rule-based systems by hand and the limitations
inherent in previous fixed and variable topology neural networks.  Unlike previous models like Gasnet \cite{husband}, chemical communication within ANT enables it to dynamically add, remove and modify modules of neurons through coarse coding. This facilitates segmentation of the search space to perform self-organized task decomposition \cite{Thangav2010}.  This also provides good scalability and generalization of sensory input \cite{Thangav2005}.  ANT is  more flexible than NEAT.  It can be initialized with a large number of neurons without the need for incremental `complexification' \cite{Thangav2010}.

As will be shown later, ANT does not rely on detailed task-specific knowledge or detailed physical models of the system. It evolves
controllers to optimize a user-specified global objective (fitness) function.
The evolutionary selection process is able to discover for itself
and exploit templates, stigmergy and mitigate the
effects of antagonism.

\section{Artificial Neural Tissue}

% \vspace{-8pt}

%\begin{figure} [h!]
%\centering
%\includegraphics[width=5.00in]{figures/architect_colour.png}
% \vspace{-12pt}
%\caption{Schematic of the Artificial Neural Tissue (ANT) Architecture. }
% \vspace{-8pt}
% \label{fig:ant}
%\end{figure}

\begin{figure} [h]
    \centering
    \includegraphics[width=6.25in,keepaspectratio,clip]{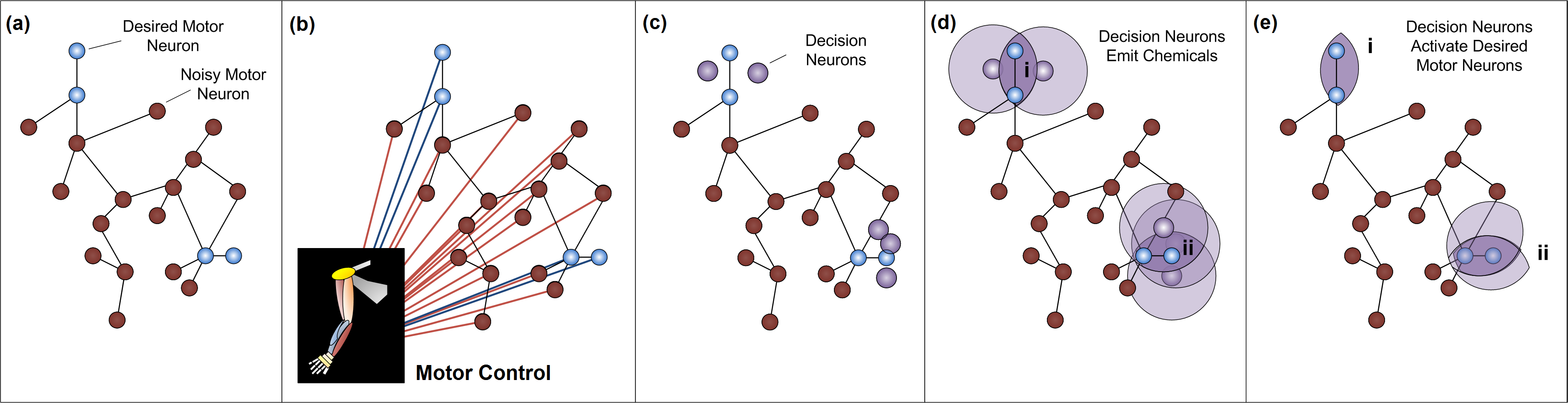}
 %\vspace{-12pt}
\caption{In a randomly generated tissue, most motor neurons would produce spurious/incoherent output (a) that would `drown out' signals from a few desired motor neurons due to spatial crosstalk \cite{jordan:jacob} (b).  This can make training intractable for difficult tasks. Neurotransmitter (chemicals) emitted by decision neurons (c) selectively activate networks of desired motor neurons in shaded regions (i) and (ii) by coarse-coding overlapping diffusion fields as shown (d).  This inhibits noisy motor neurons and eliminates spatial crosstalk (e).} \label{fig:overall}
\end{figure}

ANT \cite{Thangav2005,Thangav2008,Thangav2010} is a neural networks approach trained using evolutionary algorithms.  ANT is applied in this paper as the controller for the robot exacavator(s).  It  consists of a developmental program encoded into an artificial \emph{genome} composed of a set of genes to construct a three-dimensional artificial neural tissue. Inspired by neurobiology, ANT models chemical communication in addition to electrical communication along axons. Some neurons release chemicals that travel by diffusion and are read by other neurons, in essence serving as a `wireless' communication system to complement the `wired' one.  This chemical communication scheme is used to dynamically activate and inhibit network of neurons.  The tissue consists of two types of neural units, \emph{decision neurons} and \emph{motor-control neurons}, or simply motor neurons.   Assume a randomly generated set of motor neurons in a tissue connected by wires (Figure~\ref{fig:overall}a).  Chances are most of these neurons will produce incoherent/noisy output, while a few may produce desired functions.  If the signal from all of these neurons are  summed, then these ``noisy'' neurons would drown out the output signal  (Figure~\ref{fig:overall}b) due to spatial crosstalk \cite{jordan:jacob}.

Within ANT, decision neurons  emit chemicals that diffuse omnidirectionally shown shaded (Figure~\ref{fig:overall}d).   Coarse coding is a distributed representation that uses multiple overlapping coarse fields to encode a finer field \cite{albus,hinton}. By coarse-coding multiple overlapping diffusion fields, the desired motor neurons can be selected and spurious motor neurons inhibited. With multiple overlapping diffusion fields (Figure~\ref{fig:overall}d) shown in shaded region (ii), there is redundancy and when one decision neuron is modified (i.e. due to a deleterious mutation) the desired motor neurons are still selected.  In the following section, the computational mechanisms within the tissue is described first, followed by description of how the tissue is created.

\subsection{Motor Neurons}
\label{subsec:motor_neurons}

\begin{figure} [h]
\centering
\includegraphics[width=2.0in]{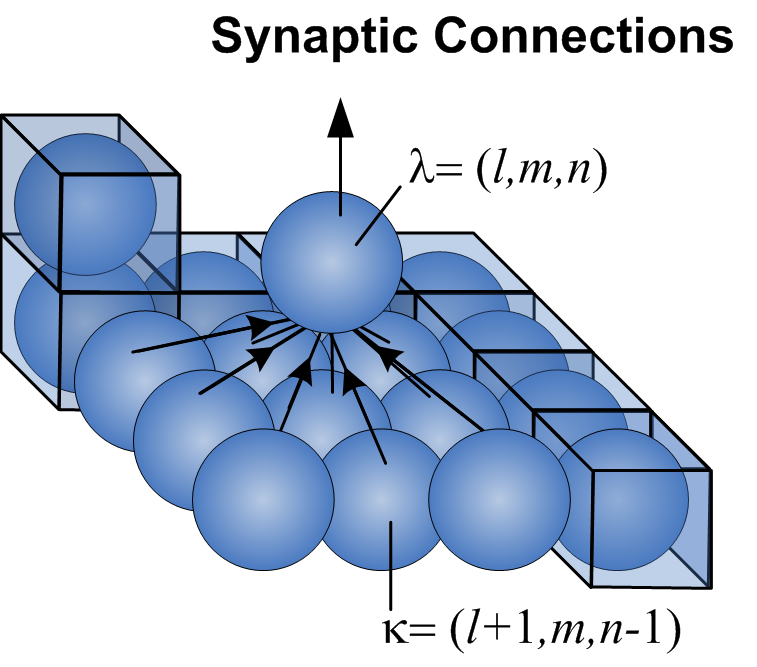}
% \vspace{-12pt}
\caption{Synaptic connections between ANT motor neurons from layer $l+1$ to $l$.}
% \vspace{-8pt}
 \label{fig:neurons.synaptic}
\end{figure}

Motor neuron, $N_{\boldlambda}$ occupies the position $\boldlambda = (l, m, n) \in
\mathbb{Z}^3$ (Figure~\ref{fig:neurons.synaptic}) and is arranged in a lattice structure.  Depending on the
activation functions used,
the state $s_{\boldlambda} \in \mathbb{S}$ of the neuron
is either binary, i.e., $\mathbb{S}_{bin} = \{0, 1\}$ or can
be real, $\mathbb{S}_p = [0, 1]$ or $\mathbb{S}_r = [-1, 1]$.

Each motor neuron $N_{\boldlambda}$  receives input from neurons
$N_{\boldkappa}$ where $\boldkappa \in\, \Uparrow\!\!(\boldlambda)$,
the nominal input set. These nominal
inputs are the $3 \times 3$ neurons centered one layer below
$N_{\boldlambda}$; in other terms, $\Uparrow\!\!(\boldlambda) =
\{(i,j,k)\,|\,i = l-1, l, l+1; \,j=m-1, m, m+1; \,k=n-1\}$.

The sensor data are represented by the activation (state) of the sensor
input neurons $N_{\boldalpha_i}, i=1 \ldots m$, summarized as $A =
\{s_{\boldalpha_1}, s_{\boldalpha_2} \ldots s_{\boldalpha_m}\}$.
The network output is represented by the
activation (state) of the output neurons $N_{\boldomega_j}, j=1 \ldots n$,
summarized as $\Omega = \{s_{\boldomega_1^1}, s_{\boldomega_2^1},s_{\boldomega_3^2}
\ldots s_{\boldomega_n^b}\}$, where $q=1 \ldots b$ specifies the
output behavior. Each output neuron commands one behavior of the
robot. (In the case of a robot, a typical behavior may be to move
forward a given distance. This may require the coordinated
action of several actuators. Alternatively, the behavior may be more
primitive such as augmenting the current of a given actuator.) If
$s_{\boldomega_j^q}=1$, output neuron $\boldomega_j$ votes to
activate behavior $q$; if $s_{\boldomega_j^q}=0$, it does not. Since
multiple neurons can have access to a behavior, an
arbitration scheme is imposed to ensure the controller is
deterministic where $p(q) = \sum_{j=1}^{n}
{\gamma(s_{\boldomega_j^i},q)s_{\boldomega_j^i}}/ n_q$ and $n_q=\sum_{j=1}^{n}{\gamma(s_{\boldomega_j^i},q)}$ is the number of output neurons connected to output behavior $q$ where $\gamma(s_{\boldomega_j^i},q)$ is evaluated as follows:
% MathType!MTEF!2!1!+-
% feaagaart1ev2aaatCvAUfeBSjuyZL2yd9gzLbvyNv2CaerbuLwBLn
% hiov2DGi1BTfMBaeXatLxBI9gBaerbd9wDYLwzYbItLDharqqtubsr
% 4rNCHbGeaGqiVu0Je9sqqrpepC0xbbL8F4rqqrFfpeea0xe9Lq-Jc9
% vqaqpepm0xbba9pwe9Q8fs0-yqaqpepae9pg0FirpepeKkFr0xfr-x
% fr-xb9adbaqaaeGaciGaaiaabeqaamaabaabaaGcbaGaeq4SdCMaai
% ikaiaadohadaWgaaWcbaGaamyyaaqabaGccaGGSaGaam4AaiaacMca
% cqGH9aqpdaGabaqaauaabeqaceaaaeaacaaIXaGaaiilaiaabccaca
% qGPbGaaeOzaiaabccacaWGIbGaeyypa0Jaam4AaiaabccacaqGGaGa
% aeiiaiaabccacaqGGaGaaeiiaaqaaiaaicdacaGGSaGaaeiiaiaab+
% gacaqG0bGaaeiAaiaabwgacaqGYbGaae4DaiaabMgacaqGZbGaaeyz
% aiaabccacaqGGaaaaaGaay5Eaaaaaa!55CB!
\begin{equation}\label{eq:check}
\gamma (s_{\boldomega_j^i} ,q) = \left\{ {\begin{array}{*{20}c}
   {1,\;\;{\mbox{if}\;\;}i = q{\;\;\;\;}}  \\
   {0,\;\;{\mbox{otherwise}\;\;}}  \\
\end{array}} \right.
\end{equation}
and resulting in behavior $q$ being activated if $p(q) \geq 0.5$.
Once the behaviors are activated they are  executed in a \emph{a
priori} sequence.

\subsection{The Decision Neuron}
\label{ch3:decision_neuron}

\begin{figure*} [h]
\centering
{\includegraphics[width=5.75in]{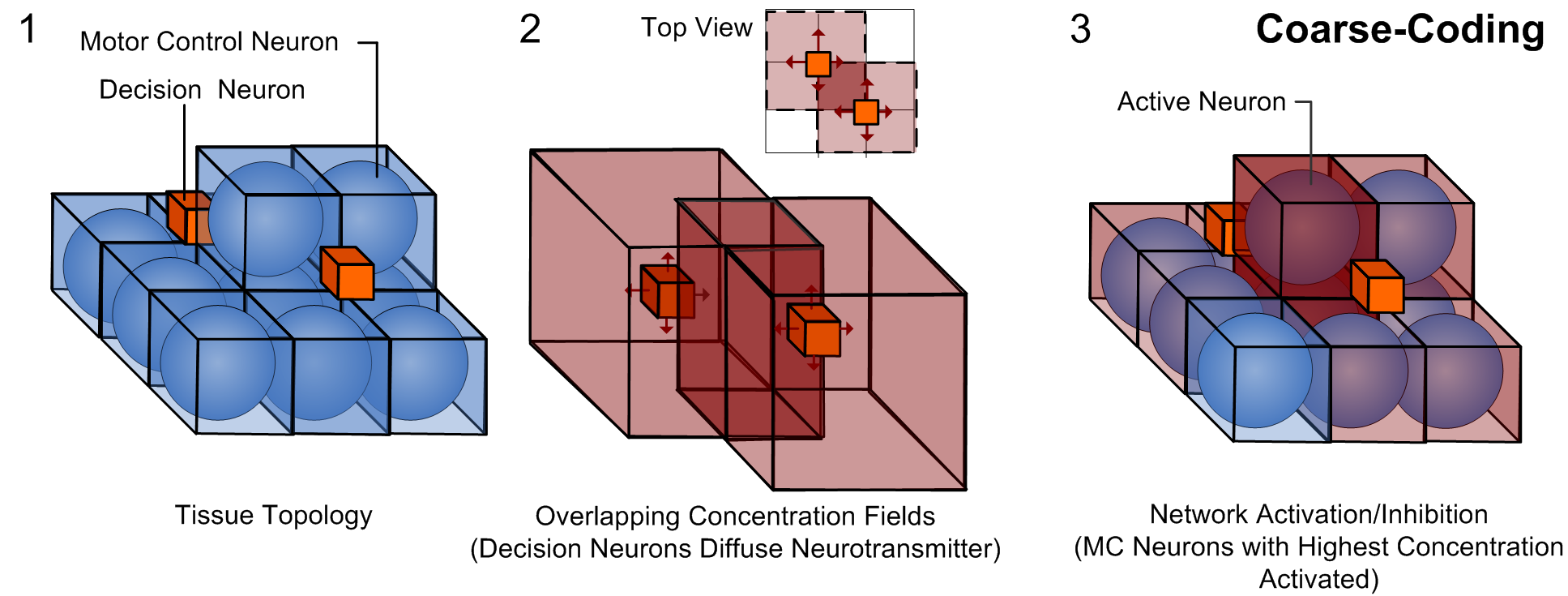}}
% \vspace{-12pt}
\caption{Coarse coding regulation being performed by two decision neurons (shown as squares) that diffuse a chemical, in turn activating a motor neuron column located at the center (right).}
\label{fig:neurons.coarsecoding}
 %\vspace{-8pt}
\end{figure*}

Decision neurons occupy nodes in the lattice as established
by their genetic parameters (Figure~\ref{fig:neurons.coarsecoding}). These neurons excite into operation or inhibit  the
motor control neurons (shown as spheres) by excreting an activation chemical. Once a motor control
neuron is excited into operation, the computation outlined in
(\ref{eq:neuron}) is performed.

Each decision neuron can be in one of two states, diffuse a neurotransmitter chemical or remain dormant. The state of
a decision neuron $T_{\boldmu}$, $s_{\boldmu}$, is binary and determined
by one of the activation functions (see Section 3.3).  The
inputs to $T_{\boldmu}$ are all the input sensor neurons
$N_{\boldalpha}$; \emph{i.e.}, $s_{\boldmu} = \psi_{\boldmu}
(s_{\boldalpha_1} \ldots s_{\boldalpha_m})$ where $\sigma_{\boldmu}
= \sum_{\boldalpha} v_{\boldalpha}^{\boldmu} s_{\boldalpha} /
\sum_{\boldalpha} s_{\boldalpha}$ and $v_{\boldalpha}^{\boldmu}$ are
the weights. The decision neuron is dormant if $s_{\boldmu} = 0$ and
releases a neurotransmitter chemical of uniform
concentration $c_{\boldmu}$ over a prescribed field of influence if
$s_{\boldmu} = 1$.

The decision neuron's field of influence is taken
to be a rectangular box extending $\pm d_{\boldmu}^r$, where $r=1,
2, 3,...$, from $\boldmu$ in the three perpendicular directions. These
three dimensions along with $\boldmu$ and $c_{\boldmu}$, the
concentration level of the virtual chemical emitted by
$T_{\boldmu}$, are encoded in the genome.

Motor control neurons within the highest chemical
concentration field are excited into operation by the decision neurons, while all others remain dormant.  Owing to the coarse coding effect,
the sums used in the weighted input of (\ref{eq:activation}) are over
only the set $\overline{\Uparrow}(\boldlambda) \subseteq\,
\Uparrow\!\!(\boldlambda)$ of active inputs to $N_{\boldlambda}$.
Likewise the output of ANT is  $\overline{\Omega}
\subseteq \Omega$.

\subsection{Activation Function}
\label{ch3:activation_functions}

Each neuron, motor and decision uses a modular activation function.  This allows selection among four possible
threshold functions of the weighted input $\sigma$.  The use of two threshold parameters allows for a
single neuron to compute the XOR function, in addition to the AND and
OR functions.  For this activation function,

\begin{equation}\label{eq:activation}
\begin{array}{rl}
    \psidown(\sigma) & = \left\{
        \begin{array}{rl}
            0, & \mbox{if}\; \sigma \geq \theta_1 \\
            1, & \mbox{otherwise}
        \end{array} \right.
    \\[4mm]
    \psiup(\sigma) &= \left\{
        \begin{array}{rl}
            0, & \mbox{if}\; \sigma \leq \theta_2 \\
            1, & \mbox{otherwise}
        \end{array} \right.
    \\[4mm]
    \psiditch(\sigma) &= \left\{
        \begin{array}{rl}
            0, & \min(\theta_1, \theta_2) \leq \sigma < \max(\theta_1, \theta_2) \\
            1, & \mbox{otherwise}
        \end{array} \right.
    \\[4mm]
    \psimound(\sigma) &= \left\{
        \begin{array}{rl}
            0, & \sigma \leq \min(\theta_1, \theta_2) \mbox{~or~} \sigma > \max(\theta_1, \theta_2) \\
            1, & \mbox{otherwise}
        \end{array} \right.
\end{array}
\end{equation}
where $\theta_1, \theta_2$ are threshold parameters and   $\theta_1, \theta_2 \in \mathbb{R}.$
\noindent The weighted input $\sigma_{\boldlambda}$  for neuron
$N_{\boldlambda}$ is nominally taken as
\begin{equation}\label{eq:neuron}
\sigma_{\boldlambda} = \frac{
    \sum_{\boldkappa \in \Uparrow(\boldlambda)} w_{\boldlambda}^{\boldkappa} s_{\boldkappa}
    }{
    \sum_{\boldkappa \in \Uparrow(\boldlambda)} s_{\boldkappa}
    }
\end{equation}
\noindent with the proviso that $\sigma = 0$ if the numerator and denominator
are zero. Also, $w_{\boldlambda}^{\boldkappa} \in \mathbb{R}$ is the
weight connecting $N_{\boldkappa}$ to $N_{\boldlambda}$. These threshold functions are summarized as
\begin{multline}\label{eq:multactivation}
\psi = (1-k_1) [ (1-k_2) \psidown + k_2 \psiup] + k_1 [(1-k_2)\psiditch + k_2 \psimound]
\end{multline}
\noindent where $k1,k2 \in {0,1}$. The activation
function is thus encoded in the genome by $k_1, k_2$ the
threshold parameters $\theta_1$ and $\theta_2$.

\subsection{Evolution and Development}
\label{ch3:evolution_and_development}

\begin{figure*} [h]
\centering
{ \includegraphics[width=6.25in]{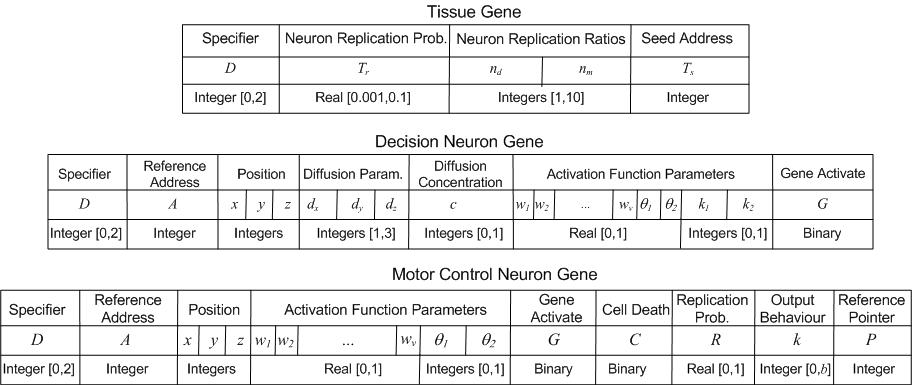}}
\caption{ANT gene map showing tissue, motor neuron and decision neuron genes.} \label{fig:genes}
\end{figure*}

A population of artificial neural tissues are evolved in an artificial Darwinian manner.  The `genome' for a tissue contains a `gene' for each cell with a specifier $D$ used to distinguish between motor control neuron, decision neuron and tissue.  A constructor protein (an autonomous program) interprets the information encoded in the gene (Figure~\ref{fig:genes}) and translates this into a cell descriptor protein.  The gene `activation' parameter is a binary flag resident in all the cell genes and is used to either express or repress the contents of the gene.  When repressed, a descriptor protein of the gene content is not created.  Otherwise, the constructor protein `grows' a cell. Each cell position is specified in reference to a seed-parent address.  A cell-death flag determines whether the cell commits suicide after being grown. Once again, this feature in the genome helps in the evolutionary process with a cell
committing suicide still occupying a volume in the lattice although it is dormant.  Evolution can decide to reinstate the cell by merely toggling a bit through mutation.

In turn mutation (manipulation of gene parameters with a uniform random distribution) to the growth program results in new cells being formed through cell division. The rate of mutation occurring on the growth program is specified for each tissue and is dependent on the cell replication probability parameter $T_r$. This probability parameter is used to determine whether a new cell is inserted.  Cell division requires a parent cell (selected with highest replication probability relative to the rest of the cells within the tissue) and copying $m\%$ of the original cell contents to a daughter cell (where $m$ is determined based on uniform random distribution).  This models a gene duplication process with the first $m\%$ being a redundant copy of an existing gene and the remaining contents being malformed in which a daughter gene adopts some functions from its parent.

The tissue gene specifies parameters that govern the overall description of the tissue.  This includes ``neuron replication probability'', a parameter that govern the probability additional cell genes are created through mutation and the ``Neuron Replication Ratio'', that determines the ratio of decision neuron genes to motor neurons genes in the tissue.  The ``Cell Type'' of each new cell is determined based on the ratio of motor control neurons to decision neurons, a parameter specified in the tissue gene.  The new neuron can be located in one of six neighboring locations (top, bottom, north, south, east, west) chosen at random and sharing a common side with the parent and not occupied by another neuron.  Furthermore a seed address is specified, that identifies the seed cell gene which will be used to construct the first cell in the tissue.

\subsection{Crossover and Mutation}
\label{ch3:crossover}

\begin{figure} [h]
    \centering
    \includegraphics[width=3.50in,keepaspectratio,clip]{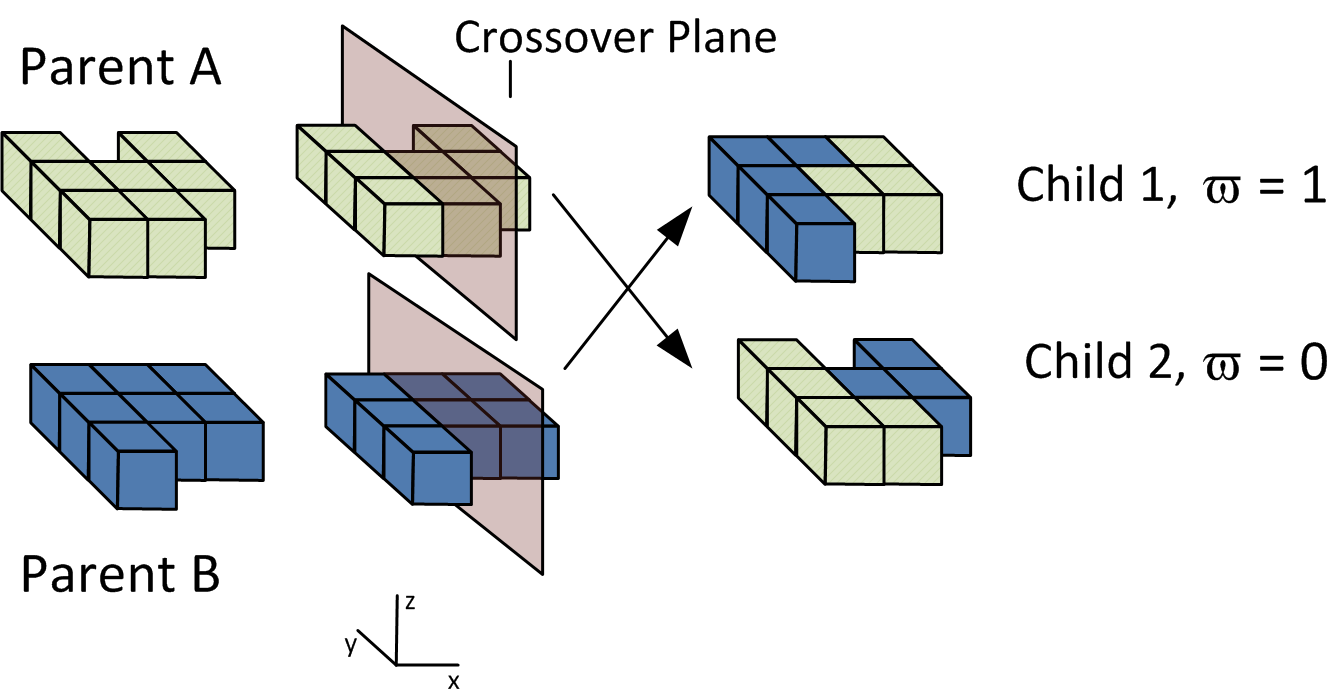}
 \vspace{-12pt}
\caption{A crossover operation between two
ANT parents. `Compatible' neuron genes are interchanged as shown resulting
in two offspring.} \label{fig:gene crossover}
 \vspace{-8pt}
\end{figure}

Within ANT, the genome is modular, with each gene defining each neuron's characteristics. Crossover  is the exchange of genes between two parents to form a child.    Before crossover, a parent affinity parameter, $\varpi \in \{0, 1\}$ is chosen at random
for each child genome.  The affinity parameter is used to establish if each
child genome has closer `affinity' to one of its parents (either parent A or parent B).  Thus if $\varpi=0$
then the genome has closer `affinity' to parent $A$ and $\varpi=1$ if it has affinity with parent $B$.
Each neuron has a unique position  $\boldlambda = (l, m, n)$ and a crossover is performed by drawing a plane (with a normal vector parallel to the {\it x} or {\it y}-axis) separating the tissue.  Cell genes on the parent
genome located on the side of the plane closer to the origin is copied directly onto
the child genome with the associated `affinity' parameter and the remaining genes
are exchanged between the parents based on a `compatibility criterion' (Figure~\ref{fig:gene crossover}).  Crossover operation does not result in arbitrary separation and exchange of gene contents.

The `compatibility criterion' imposes the following condition, that the gene for neuron $N_{\boldlambda_1}$ from parent $A$ and $N_{\boldlambda_2}$ from parent $B$ could be exchanged  if $\boldlambda_1 = \boldlambda_2$, i.e., have the same position after development and only when \emph{both} genes are expressed or repressed during development. Thus child 1 with $\varpi =0$ (affinity to parent $A$) assumes the gene for ${}^{B}N_{\boldlambda_2}$ and child 2 with $\varpi =1$ (affinity to parent $B$) assumes ${}^{A}N_{\boldlambda_1}$.  If the compatibility criterion is not met, then no exchange occurs, and thus ${}^{A}N_{\boldlambda_1}$ is passed onto child 1 and ${}^{B}N_{\boldlambda_2}$ is passed onto child 2.
If ${}^{A}N_{\boldlambda_1}$ is not expressed in parent $A$ and ${}^{B}N_{\boldlambda_1}$ is expressed in parent $B$,  then this pair of genes fail the `compatibility criterion.'

A detailed description and analysis of how this evolutionary approach utilizing coarse coding leads to task decomposition of complex task and discovery of novel behavior is presented in Section 5.2.

%%%%%%%%%%%%%%%%%%%%%%%%%%%%%%%%%%%%%%%%%%%%%%%%%%%%%%%%%%%%%%%%%%%%%%%%%%%%%%%%
\section{Excavation Task}
\label{sec:simulation}

\begin{figure*} [h]
\centering
        \includegraphics[width=6.25in]{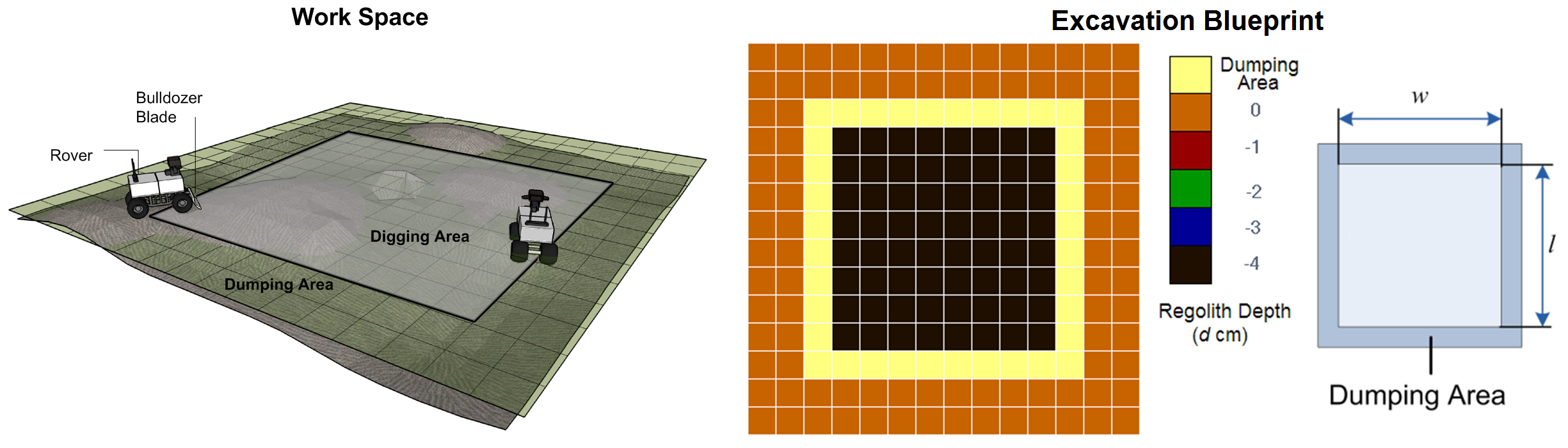}
\caption{A typical workspace for the excavation task (left). A corresponding excavation blueprint consisting of an excavation area surrounded by a dumping area (right).  The blue print is compared against the current workspace topology to compute a fitness that range from 0 to 1, where 1 corresponds to the desired topology as specified by the blueprint.  }
\label{fig:explain}
\end{figure*}

The excavation task is intended to demonstrate the feasibility of
autonomous teams of robots digging pits and clearing landing pads for lunar base construction.  In these experiments teams of robots are equipped with bulldozer blades that are used to push regolith. Self-organized task decomposition may be needed to accomplish the task
given a global fitness function that does not give instructions on how to solve the task or bias for a particular solution strategy.  A typical training environment is shown in Figure~\ref{fig:explain}.  The
workspace is modeled as a two-dimensional grid
with each robot occupying four grid squares.  For this task, the controller needs
to possess several capabilities to complete the task, including interpreting excavation
blueprints, performing layered digging and avoid burying or trapping
other robots.  The blueprint defines the location of the
dumping area and target depth of the excavation area (Figure~\ref{fig:explain}). Each robot
controller has access only to a local zone within the excavation blueprint at a time. Much like how insects have hard coded genes to sense templates in their environment, the robots have pre-programmed capability (similar to hard coded genes) to sense the various states of the goal map.   However this is insufficient in completing an excavation task and these various states need to be correctly interpreted to perform the correct actions. The
fitness function $f$  for the task is given as follows:

% MathType!MTEF!2!1!+-
% feaagaart1ev2aqatCvAUfeBSjuyZL2yd9gzLbvyNv2CaerbuLwBLn
% hiov2DGi1BTfMBaeXatLxBI9gBaerbd9wDYLwzYbItLDharqqtubsr
% 4rNCHbGeaGqiVu0Je9sqqrpepC0xbbL8F4rqqrFfpeea0xe9Lq-Jc9
% vqaqpepm0xbba9pwe9Q8fs0-yqaqpepae9pg0FirpepeKkFr0xfr-x
% fr-xb9adbaqaaeGaciGaaiaabeqaamaabaabaaGcbaGaamOzaiabg2
% da9maalaaabaWaaabmaeaadaaeWaqaaiaadchadaWgaaWcbaGaamyA
% aiaacYcacaWGQbaabeaakiabgwSixlaadwgadaahaaWcbeqaaiabgk
% HiTiaaikdadaabdaqaaiaadEgadaWgaaadbaGaamyAaiaacYcacaWG
% QbaabeaaliabgkHiTiaadIgadaWgaaadbaGaamyAaiaacYcacaWGQb
% aabeaaaSGaay5bSlaawIa7aaaaaeaacaWGPbGaeyypa0JaaGymaaqa
% aiaadMeaa0GaeyyeIuoaaSqaaiaadQgacqGH9aqpcaaIXaaabaGaam
% OsaaqdcqGHris5aaGcbaWaaabmaeaadaaeWaqaaiaadchadaWgaaWc
% baGaamyAaiaacYcacaWGQbaabeaaaeaacaWGPbGaeyypa0JaaGymaa
% qaaiaadMeaa0GaeyyeIuoaaSqaaiaadQgacqGH9aqpcaaIXaaabaGa
% amOsaaqdcqGHris5aaaaaaa!65C5!
\begin{equation}\label{eq:excavation_fitness}
f = \frac{{\sum\nolimits_{j = 1}^J {\sum\nolimits_{i = 1}^I {\vartheta_{i,j}
\cdot e^{ - 2\left| {g_{i,j}  - z_{i,j} } \right|} } }
}}{{\sum\nolimits_{j = 1}^J {\sum\nolimits_{i = 1}^I {\vartheta _{i,j} } } }}
\end{equation}

\noindent where $I$ and $J$ are the dimensions of the entire area and $\sum\nolimits_{j
= 1}^J {\sum\nolimits_{i = 1}^I {\vartheta _{i,j} } } > 0$ and $\vartheta _{i,j} = 1$ if grid square $(i,j)$ is to be excavated and 0 otherwise;  $g_{i,j}$ is the target depth and $z_{i,j}$  is the current depth. This objective function is used to train the robot controllers for the task at hand. The function produces a real value typically that can range from 0 to 1, where 1 indicates the current excavation area matches the blue print topology and 0 when it doesn't match the topology.  This fitness function maybe used as a quantitative metric to compare the performance of various excavation systems.

\subsection{Robot Model}
\label{sec:robot_model}
\begin{figure*} [h]
\centering
        \includegraphics[width=6.25in]{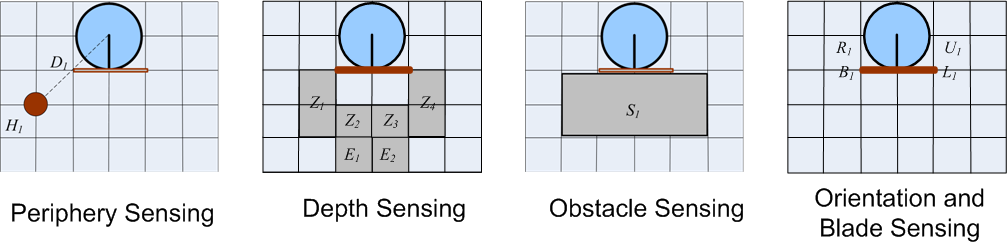}
\caption{Robot input sensor mapping for the simulation model.}
\label{fig:input_map}
\end{figure*}

For these experiments, the inputs to controllers evolved using the ANT methodology (referred to as ANT controllers) are shown in Table~\ref{tbl:sensor_inputs_digging} and their location is shown in Figure~\ref{fig:input_map}. The robots have access to current position $(x,y)$ from localization scans performed in simulation. The variable $z$  is computed through estimated integration of changes in depth values. The discretized $x$ and $y$ coordinates are used
to look up the goal depth $g_{x,y}$ from the excavation blueprint of each grid square region in front of the robot.

All raw sensory input data are discretized and fed to the controller. $Z_1$\ldots$Z_4$ and $E_1$\ldots$E_2$ are obtained using simulated ground scans of the grid squares shown.  For practical implementation these ground scans would be obtained from a LIDAR or TRIDAR system.  $E_1$\ldots$E_2$ are obtained by comparing the soil depth in front of the rover to the wheel depth.  A simulated sensor, $S_1$ is used to  detect obstacles at the front. Robot tilt, heading of nearest robot, distance of nearest robot and whether a robot is stuck or not are represented using $R_1$, $H_1$, $D_1$ and $U_1$ respectively. The heading and distance to the nearest robot are used because front obstacle detection alone may not sufficient to detect obstacle due to inherent blind spots. It is also important for the robot to determine whether it is stuck or not after executing its move behavior. This can be due to any number of reasons and provides the opportunity for the controller to react.  In addition, the state of the attached bulldozer blade is provided.  The bulldozer blade can be in one of 4 positions, above ground,  level, below ground and home position.   This enables a robot to fill, push, dig or traverse over the soil respectively. In addition, simulated force sensors are mounted to the blade and measures the net axial force against the blade.  The robots can push a maximum of 24 units of soil and this is rescaled to between 0 and 4 for the blade force load, $L_1$.  The robots also have access to one memory bit ($M_1$) that can be manipulated using several basis behaviors. Together there are $5^4 \times 3^2 \times 4 \times 5 \times 2 \times 4 \times 4 \times 2^3=8.6 \times 10^7$ possible combination of sensor inputs.

\begin{table*}[t!]
\caption{Excavation: Sensor Inputs} \label{tbl:sensor_inputs_digging}
\centering
\footnotesize{
\begin{tabular}{c l l}
\hline
{\bf Sensor Variables}& {\bf Function}      & {\bf Description} \\
\hline\hline
$Z_1 \ldots Z_4$    & Depth Sensing Relative to Goal Depth   & Level, Above, Below, Don't Care, Dump  \\
$E_1, E_2$        & Depth Sensing Relative to Ground       & Above, Below or Level                  \\
$B_1$               & Blade Position                         & Below, Level, Above, Home              \\
$L_1$              & Blade Force Sensor                      & $0-4$                                  \\
$S_1$               & Front Obstacle Detection               & Obstacle, No Obstacle                  \\
$D_1$               & Separation Distance From Nearest Robot & $0-3$                                  \\
$H_1$               & Heading From Nearest Robot             & North, East, West, South               \\
$R_1$              & Robot Tilted Downwards                 & True, False                            \\
$U_1$              & Robot Stuck                            & True, False                            \\
$M_1$              & Memory Variable                        & 0, 1                                   \\

\hline
\end{tabular}}
\end{table*}

Table~\ref{tb:excavate_behaviors} lists the basis behaviors the robot can perform sequentially according to the given order within a single timestep.  These basis behaviors are generic and can be used for many different tasks.  It is up to the ANT controller to determine when to execute these basis behaviors (using sensory input) to solve the overall task.  Furthermore, the excavation behaviors allow for actuation of a bulldozer blade.  The blade can be raised or lowered to one of four positions as stated earlier.

\begin{table*}[t!]
\caption{Excavation Basis Behaviors} \label{tb:excavate_behaviors}
 \vspace{2pt}
\centering
\footnotesize{
\begin{tabular}{c l l}
\hline
{\bf Order}           & {\bf Behavior}      & {\bf Description}                                                \\
\hline\hline
1               & Throttle Up   & Set drive system to high power mode from default.                       \\
2               & Move Forward  & Move one grid square forward                               \\
3               & Move Backward & Move one grid square backward                               \\
4               & Random Turn   & Randomly turn $90\degreeb$ right or left.                  \\
5               & Turn Right    & Turn $90\degreeb$ right                                    \\
6               & Turn Left     & Turn $90\degreeb$ left                                     \\
7               & Blade Above   & Set blade above ground $d$ cm                              \\
8               & Blade Below   & Set blade below ground $d$ cm                              \\
9               & Blade Level   & Set blade level to ground $d$ cm                           \\
10              & Blade Home    & Retract blade to home position (no contact with regolith). \\
11              & Bit Set       & Set memory bit $1$ to 1                                    \\
12              & Bit Clear     & Set memory bit $1$ to 0                                    \\
\hline
\end{tabular}}
\end{table*}

\subsection{Robot Excavation Model}

The  workspace is discretized into a two-dimensional grid world, with each grid square having dimensions $l_x \times l_y$. In this model the robots are equipped with a bulldozer blade which is used to dig and push regolith. The robot as explained earlier can only move forward and backward along cardinal directions and the new position after these behaviors is $(x_i + \Delta x,y_i + \Delta y)$ at time $t + \Delta t$ where $\Delta x,\Delta y \in \mathbb{Z}$ and  $\Delta t$ is the timestep required to execute the behavior.  In addition, $\left| {\Delta x} \right| + \left| {\Delta y} \right| = 1$.  A simplified soil interaction model is used to simulate excavation in the controller training environment.  The simulation environment is used to track the soil height for each grid square, where $h(i,j,t)$ is the soil height at grid square $(i,j)$ at time $t$.  In the training environment, the robot occupies four grid squares.  The digging apparatus, which is a bulldozer blade occupies two grid squares next to each other and are at location $(x_1,y_1)$ and $(x_2,y_2)$.

The soil interaction model assumes the soil is incompressible and evenly distributed within each grid square. It is further assumed that the interaction between the soil is through a bulldozer blade that is used to push or fill soil according to the behaviors in Table~\ref{tb:excavate_behaviors}.  The bulldozer blade, $B_1$ can interact with the soil when its in one of three states, Below, Level and Above ground.   This interaction with the soil is computed as follows:

% MathType!MTEF!2!1!+-
% feaagCart1ev2aaatCvAUfeBSjuyZL2yd9gzLbvyNv2CaerbuLwBLn
% hiov2DGi1BTfMBaeXatLxBI9gBaerbd9wDYLwzYbItLDharqqtubsr
% 4rNCHbGeaGqiVu0Je9sqqrpepC0xbbL8F4rqqrFfpeea0xe9Lq-Jc9
% vqaqpepm0xbba9pwe9Q8fs0-yqaqpepae9pg0FirpepeKkFr0xfr-x
% fr-xb9adbaqaaeGaciGaaiaabeqaamaabaabaaGcbaGaamiAaiaacI
% cacaWG4bGaey4kaSIaeyiLdqKaamiEaiaacYcacaWG5bGaey4kaSIa
% eyiLdqKaamyEaiaacYcacaWG0bGaey4kaSIaeyiLdqKaamiDaiaacM
% cacqGH9aqpcaWGObGaaiikaiaadIhacqGHRaWkcqGHuoarcaWG4bGa
% aiilaiaadMhacqGHRaWkcqGHuoarcaWG5bGaaiilaiaadshacaGGPa
% Gaey4kaSIaamiAaiaacIcacaWG4bGaaiilaiaadMhacaGGSaGaamiD
% aiaacMcacqGHsislcaWG6bWaaSbaaSqaaiaadEhaaeqaaOGaeyOeI0
% IaamOyamaaBaaaleaacaWGObaabeaakiabgwSixlabew7aLjaacIca
% caWGwbWaaSbaaSqaaiaadkgacaWGSbGaamyyaiaadsgacaWGLbaabe
% aakiaacMcaaaa!6D07!

\begin{multline}\label{eq:excavation_fitness}
h({x_i},{y_i},t + \Delta t) = h({{\tilde x}_i} + \Delta x,{{\tilde y}_i} + \Delta y,t) + h({{\tilde x}_i},{{\tilde y}_i},t) - {z_{wi}} - {b_h} \cdot \varepsilon ({V_{blade},b_h})
\end{multline}

\noindent where, $i= [1,2]$, $({x_i},{y_i})$ are the current positions and $({\tilde x_i},{\tilde y_i})$ are the old positions of the blade.  $z_{wi}$ is the depth of front wheel $i$ and $b_h$ is blade position, where $b_h \in {-1,0,1} $ and $V_{blade} \geq 0$ and is the volume of soil in front of the blade.  After the soil height against the blade is updated, the soil height underneath the front wheel (old position of the blade) is updated as follows:
% MathType!MTEF!2!1!+-
% feaagCart1ev2aaatCvAUfeBSjuyZL2yd9gzLbvyNv2CaerbuLwBLn
% hiov2DGi1BTfMBaeXatLxBI9gBaerbd9wDYLwzYbItLDharqqtubsr
% 4rNCHbGeaGqiVu0Je9sqqrpepC0xbbL8F4rqqrFfpeea0xe9Lq-Jc9
% vqaqpepm0xbba9pwe9Q8fs0-yqaqpepae9pg0FirpepeKkFr0xfr-x
% fr-xb9adbaqaaeGaciGaaiaabeqaamaabaabaaGcbaGaamiAaiaacI
% cacaWG4bWaaSbaaSqaaiaadMgaaeqaaOGaaiilaiaadMhadaWgaaWc
% baGaamyAaaqabaGccaGGSaGaamiDaiabgUcaRiabgs5aejaadshaca
% GGPaGaeyypa0JaamOEamaaBaaaleaacaWG3bGaamyAaaqabaGccqGH
% RaWkcaWGIbWaaSbaaSqaaiaadIgaaeqaaOGaeyyXICTaeqyTduMaai
% ikaiaadAfadaWgaaWcbaGaamOyaiaadYgacaWGHbGaamizaiaadwga
% aeqaaOGaaiykaaaa!5415!
\begin{equation}\label{eq:excavation_fitness}
h({{\tilde x}_i},{{\tilde y}_i},t + \Delta t) = {z_{wi}} + {b_h} \cdot \varepsilon ({V_{blade},b_h})
\end{equation}
\noindent where $\varepsilon ({V_{blade},b_h})$ is given as follows:

% MathType!MTEF!2!1!+-
% feaagCart1ev2aaatCvAUfeBSjuyZL2yd9gzLbvyNv2CaerbuLwBLn
% hiov2DGi1BTfMBaeXatLxBI9gBaerbd9wDYLwzYbItLDharqqtubsr
% 4rNCHbGeaGqiVu0Je9sqqrpepC0xbbL8F4rqqrFfpeea0xe9Lq-Jc9
% vqaqpepm0xbba9pwe9Q8fs0-yqaqpepae9pg0FirpepeKkFr0xfr-x
% fr-xb9adbaqaaeGaciGaaiaabeqaamaabaabaaGcbaGaeqyTduMaey
% ypa0ZaaiqaaeaafaqabeGabaaabaGaaGimaiaacYcacaqGGaGaaeyA
% aiaabAgacaqGGaGaamOvamaaBaaaleaacaWGIbGaamiBaiaadggaca
% WGKbGaamyzaaqabaGccqGH9aqpcaaIWaGaaeiiaiaabggacaqGUbGa
% aeizaiaabccacaWGIbWaaSbaaSqaaiaadIgaaeqaaOGaeyyzImRaaG
% imaaqaaiaaigdacaGGSaGaaeiiaiaab+gacaqG0bGaaeiAaiaabwga
% caqGYbGaae4DaiaabMgacaqGZbGaaeyzaiaabccacaqGGaGaaeiiai
% aabccacaqGGaGaaeiiaiaabccacaqGGaGaaeiiaiaabccacaqGGaGa
% aeiiaiaabccacaqGGaGaaeiiaiaabccacaqGGaGaaeiiaiaabccaaa
% aacaGL7baaaaa!64D7!
\begin{equation}\label{eq:excavation_fitness}
\varepsilon  = \left\{ {\begin{array}{*{20}{c}}
{0,\;{\rm{ if }}\;{V_{blade}} = 0\;{\rm{ and }}\;{b_h} \ge 0}\\
{1,\;{\rm{ otherwise\;\;\;\;\;\;\;\;\;\;\;\;\;\;\;\;\;\;\;\;}}}
\end{array}} \right.
\end{equation}

\noindent The volume of soil in front of the blade, $V_{blade}$ is given as follows:
% MathType!MTEF!2!1!+-
% feaagCart1ev2aaatCvAUfeBSjuyZL2yd9gzLbvyNv2CaerbuLwBLn
% hiov2DGi1BTfMBaeXatLxBI9gBaerbd9wDYLwzYbItLDharqqtubsr
% 4rNCHbGeaGqiVu0Je9sqqrpepC0xbbL8F4rqqrFfpeea0xe9Lq-Jc9
% vqaqpepm0xbba9pwe9Q8fs0-yqaqpepae9pg0FirpepeKkFr0xfr-x
% fr-xb9adbaqaaeGaciGaaiaabeqaamaabaabaaGcbaGaamOvamaaBa
% aaleaacaWGIbGaamiBaiaadggacaWGKbGaamyzaaqabaGccqGH9aqp
% daWadaqaaiaadIgacaGGOaGaamiEamaaBaaaleaacaWGPbaabeaaki
% aacYcacaWG5bWaaSbaaSqaaiaadMgaaeqaaOGaaiykaiabgkHiTiaa
% dQhadaWgaaWcbaGaam4DaiaadMgaaeqaaOGaeyOeI0IaamOyamaaBa
% aaleaacaWGObaabeaaaOGaay5waiaaw2faaiabgwSixlaadYgadaWg
% aaWcbaGaamiEaaqabaGccqGHflY1caWGSbWaaSbaaSqaaiaadMhaae
% qaaaaa!559D!
\begin{equation}\label{eq:excavation_fitness}
{V_{blade}} = \sum\limits_{i = 1}^2 {\left[ {h({x_i},{y_i}) - {z_{wi}} - {b_h}} \right] \cdot {l_x} \cdot {l_y}}
\end{equation}
where $l_x$ and $l_y$ are the length and width of each grid square. When the blade is positioned below wheel depth, this results in scraping and accumulation of soil in front of the blade.  When the blade is level, this results in pushing of already accumulated soil along the terrain and when the blade is positioned above ground, the accumulated soil is dislodged.

\subsection{Training}

Noting that each behavior in Table~\ref{tb:excavate_behaviors} can be triggered or not for any one of $8.6 \times 10^7$ possible combination of sensor inputs, there is a total of $2^{12\times 8.6 \times 10^7} \approx 10^{3\times 10^8}$ possible states in the search space! Task decomposition is often necessary to tackle very large search spaces and find desired solutions. ANT using its coarse-coding scheme described earlier is shown to perform task decomposition \cite{Thangav2010} and is a good candidate to tackle this excavation task with its large task space. ANT controllers are first trained (evolved) in this simplified training environment.  The genome for ANT is shown in  Figure~\ref{fig:genes}.

Darwinian selection is performed using the given fitness function averaged over 100 different scenarios \cite{goldberg}.  The population is randomly generated with an initial number of neurons ranging from 40 to 120 neurons.  The neurons are feed forward and the second layer of neurons is fully connected to all the sensory input neurons.  Furthermore, the ANT growth program is restricted to a maximum of four layers in total.  The tissue however can grow in area.  The fitness $f$ for each controller is calculated for an excavation area, dug to a goal depth of $d$ below ground and area spanning $l \times w $ squares.  The worksite area is $8 \times  8$ squares and a goal depth of $d =$ 1, 2 or 3 units below ground.  The robots are initialized at random positions on the worksite.The population size for training is $P=100$, with crossover probability $p_c=0.7$, mutation probability $p_m=0.025$ and a tournament size of $0.06P$. The fitness of the fittest individual from the population during each generation is taken to be the system fitness.  Each training runs lasts 5,000 generations and is repeated 30 times to obtain the average system fitness.

\section{Simulation Results}

\begin{figure} [h]
    \centering
    \includegraphics[width=4in]{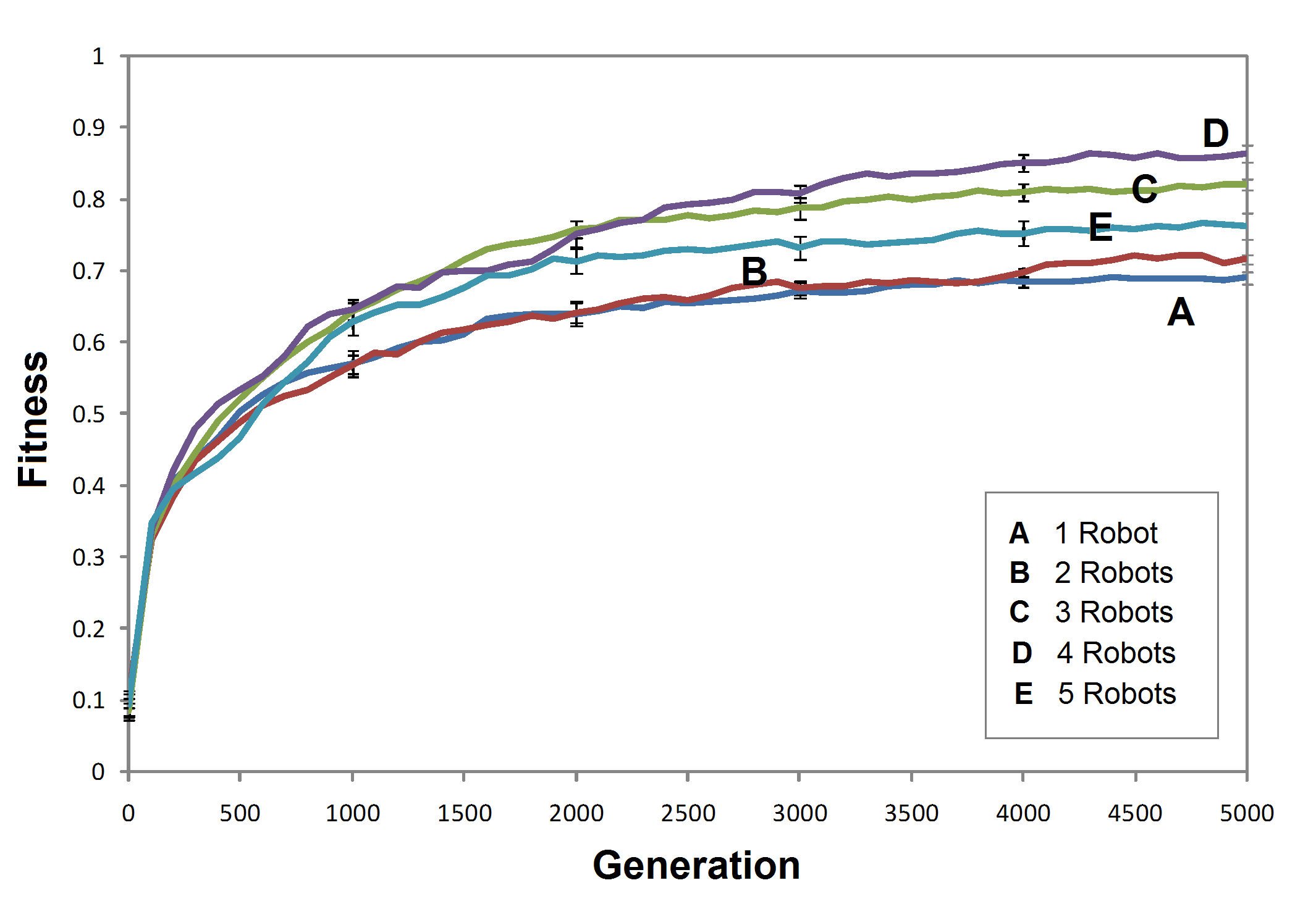}
 \vspace{-12pt}
\caption{Fitness comparison of ANT based solutions during training,
for between 1 and 5 robots averaged over 30 runs.} \label{fig:ex_evo}
 \vspace{-5pt}
\end{figure}

Figure~\ref{fig:ex_evo} shows the population best fitness  of the
system evaluated at each generation of the artificial
evolutionary run. The system performance is
affected by the number of robots per digging area ($8 \times  8$
squares).   A single robot is not as efficient as 4 robots working in parallel with each robot having a
smaller area to cover. As will be shown later in Section 5.3, with more than 4 robots for a $8 \times  8$ area, the problem of antagonism arises when multiple robots trying to perform the same task interfere with one another and reduce the overall
efficiency of the group.

\begin{figure} [h]
    \centering
    \includegraphics[width=4in]{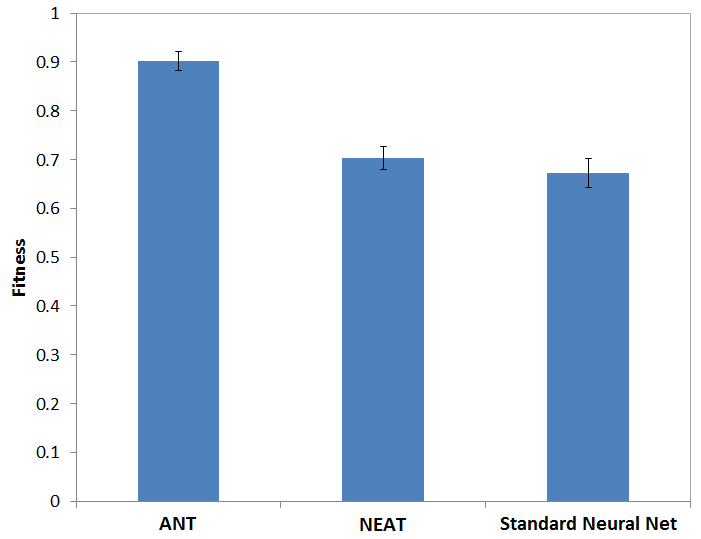}
\vspace{-12pt}
\caption{Maximum fitness averaged over 30 training runs for ANT, NEAT and standard neural networks.} \label{fig:neural_net_comparison}
\vspace{-5pt}
\end{figure}

First we consider the effect of topology on controller training performance. The performance of standard fixed topology neural networks and NEAT \cite{kstanley}, a variable topology neural network methodology is compared in Figure~\ref{fig:neural_net_comparison}.  For this comparison, topologies for NEAT and standard neural networks are randomly generated and contain between 40 and 120 neurons.  For the fixed neural networks, the neurons are feed forward and like ANT, restricted to a maximum of four layers that includes the sensor input layer and output layer. The transfer function used is the modular activation function.

\begin{figure} [h]
    \centering
    \includegraphics[width=4in]{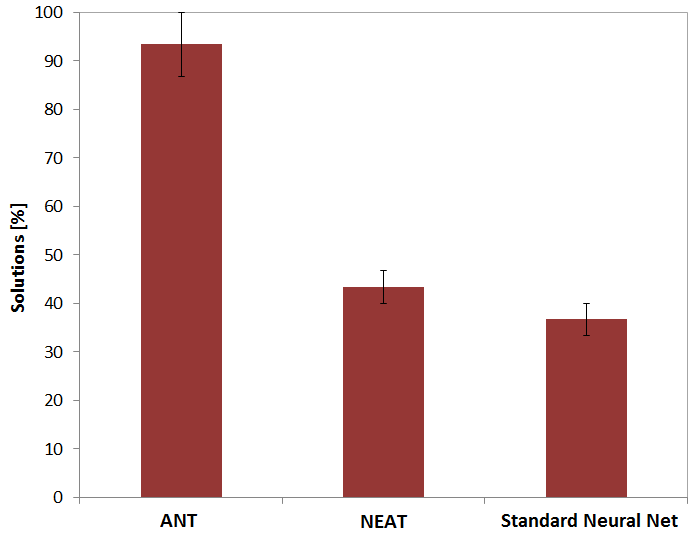}
 %\vspace{-12pt}
\caption{Probability of finding a solution (fitness $\geq 0.9$) averaged over 30 training runs for ANT, NEAT and standard neural networks.} \label{fig:neural_net_comparison2}
 %\vspace{-5pt}
\end{figure}

ANT shows nearly a 30\% improvement in fitness performance over the best of these conventional approaches.  Importantly, ANT obtains a  fit solution (fitness 0.9 or higher) with nearly a two-folds advantage over NEAT \cite{kstanley} (Figure~\ref{fig:neural_net_comparison2}). In both cases, ANT shows a substantial improvement over standard neural networks.  Further analysis of the robustness and scalability of conventional neural networks and ANT is presented in Section 5.3.  These results show that ANT performs substantially better in terms of scalability and robustness than conventional neural networks.
\begin{figure} [h]
    \centering
    \includegraphics[width=4in]{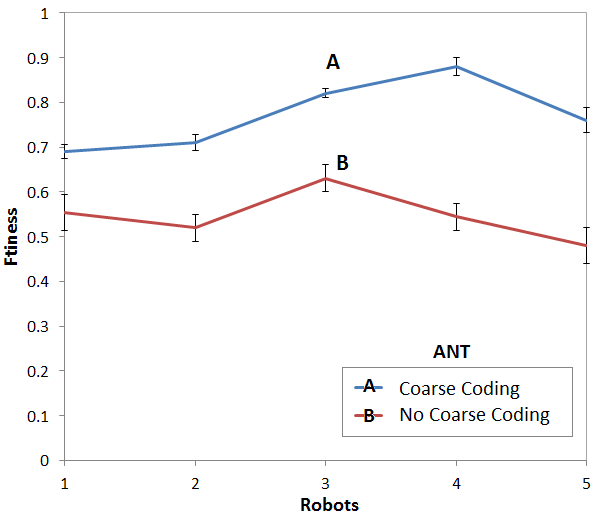}
 %\vspace{-12pt}
\caption{Maximum fitness of ANT with and without-coarse coding for between 1 and 5 robots.} \label{fig:coarse_coding_comp}
 %\vspace{-5pt}
\end{figure}

In conventional fixed and variable topology networks there
tends to be more `active' synaptic connections present (since all
neurons are active), and thus takes longer for each neuron to
tune these connections to the sensory inputs. ANT is an improvement as the topology is evolved and decision neurons learn to mask out spurious neurons.  The net result is that ANT produces fitter solutions in fewer genetic evaluations compared to conventional neural networks.

Further comparisons are performed to determine the key features within ANT that contribute to its improved performance over conventional neural networks.  In this study, the coarse-coding functionality is turned off, to determine its contribution to ANT's overall performance.  The results show a significant drop in performance (Figure~\ref{fig:coarse_coding_comp}).  The overall performance without coarse-coding is comparable to a variable topology architecture such as NEAT.  This confirms that it is the coarse-coding functionality that provides ANT its advantage over conventional neural networks.

\subsection{Evolution of Behaviors through Task Decomposition}

In this section we present evidence suggesting how self-organized
task decomposition occurs in ANT controllers. With human devised task decomposition,
the first step is to devise the task objective and then figure out the necessary subtasks, followed by further partitioning of the subtasks until all components of the task are identified and readily solvable. With self-organized task decomposition, there is no supervisor to break up the objective function into the necessary subtasks.  Instead modules form that without any central coordination and using local information learn to solve certain subtasks through a process of trial and error as will be shown here. These modules in turn interact and cooperate to solve the overall task.

\begin{figure} [h]
    \centering
    \includegraphics[width=6.25in]{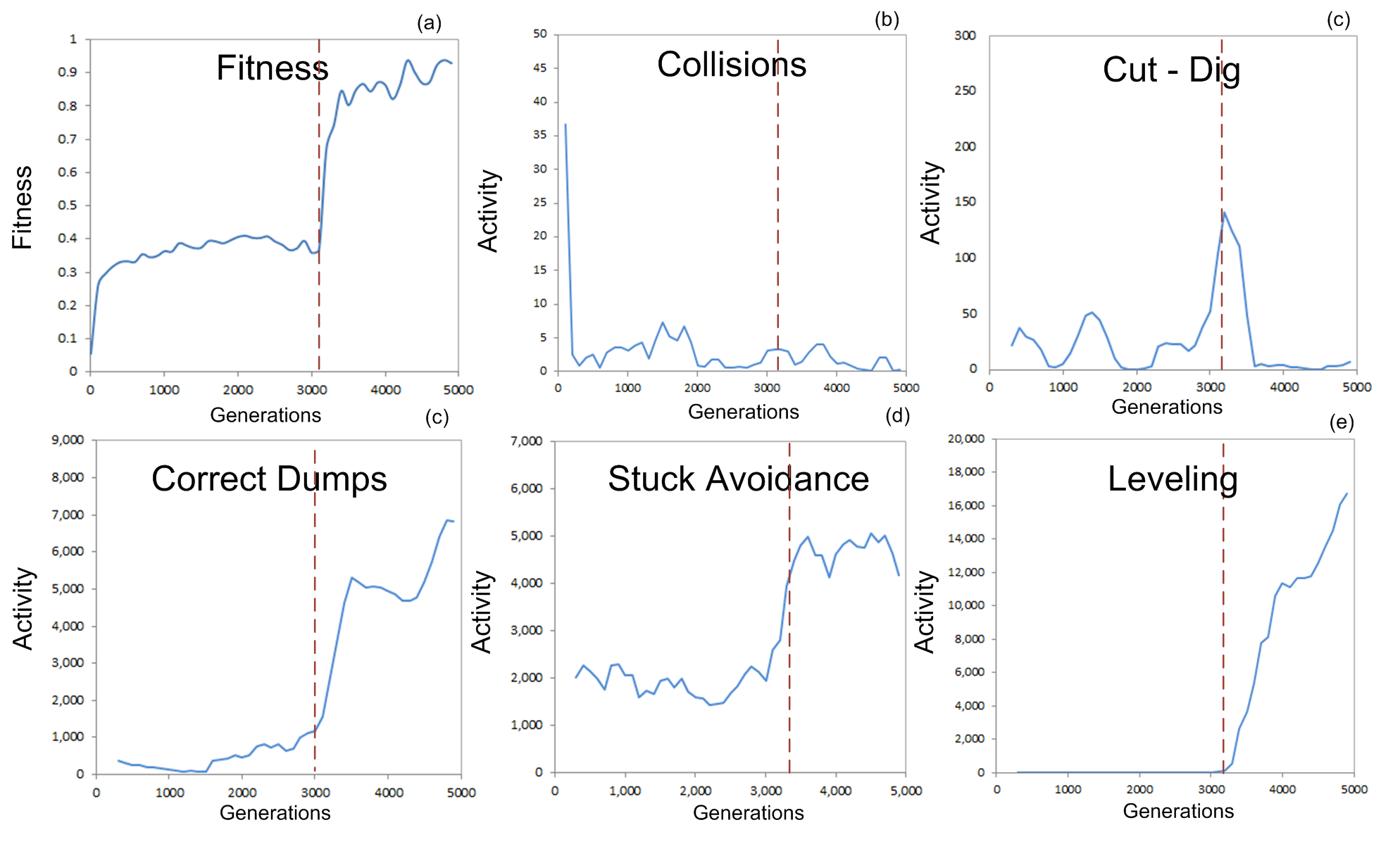}
 \vspace{-12pt}
\caption{(a) Fitness (b)-(f) Evolution of important behaviors for the excavation task during a typical evolutionary training run of ANT. At generation 3,100 new behaviors are discovered foreshadowing a rapid rise in fitness and convergence to a solution.} \label{fig:task_evo}
 \vspace{-5pt}
\end{figure}

Figure~\ref{fig:task_evo} shows a typical ANT training run for the excavation task.  Statistics are obtained of key sensor-behavior combinations listed in Table~\ref{tbl:detect} during a typical training run.  In this scenario, 4 robots are used, using the standard environment described in Section 4.3.  Figure~\ref{fig:task_evo}a shows the controllers converge to a solution through a series of punctuated rises in several identified behaviors. These results give insight into how ANT solves the excavation task. The results suggests that the controllers evolve certain critical behaviors such as simple obstacle avoidance early during evolutionary training  (Figure~\ref{fig:task_evo}b).  This is followed by related behaviors, such as stuck avoidance (Figure~\ref{fig:task_evo}e) (i.e. avoiding getting stuck in soil). These behaviors are a requirement for the controllers to effectively move  around the experiment area to perform excavation. Next, the controllers evolve to steadily improve the accuracy of performing correct dumping behaviors (Figure~\ref{fig:task_evo}d).

At around 3,100 generation, a sharp rise in fitness is observed (Figure~\ref{fig:task_evo}a), rapidly converging towards a fitness of 0.9 (solution to the task). The controllers evolve to find critical behaviors to solve the task through a process of trial and error, after evolving prerequisite behaviors such as obstacle avoidance \cite{Thangav2010}. This evidence of performing trial and error to improve fitness can be seen through increased activity of cut-dig which enables the controllers to randomly obtain a fitness advantage over rest of the population (Figure~\ref{fig:task_evo}c).  Cut-dig requires the blade be positioned below ground to perform a sharp incision cut into the soil.   As the robot continues moving forward, a discretized sloped cut is obtained instead of the required level cut.  Therefore a series of these cut-dig behaviors and back-fill behaviours are switched on and off for short distances to approximate a level cut.
	
This combined with improved dumping performance foreshadows improvement in fitness (Figure~\ref{fig:task_evo} a,c,d).  Although this digging approach is not  every efficient, it increases fitness pressure to find more efficient ways.  This results in the appearance of fine tuned digging behaviors that require fewer, inefficient incision cut-dig behaviors and more level digging.  Level digging requires the robot first perform an incision cut and then continue pushing soil at a level depth.
	
Thus with this improvement, the cut-dig behavior doesn't have to be used as often and can match blueprints with better accuracy (Figure~\ref{fig:task_evo}c).  The cut-dig behavior acts as a scaffolding mechanism, that is used at a critical time to evolve efficient digging behaviors. However for the controllers to achieve a high fitness several other behaviors are acquired.  Once the controllers achieve a high accuracy in following the excavation blueprint and performing digging and correct dumping behavior, this results in a significant improvement in fitness.  Further refinement occurs with the appearance of leveling behaviors (Figure~\ref{fig:task_evo}f) that enables the robots to wander and correct small mistakes.  This process of wandering and correcting mistakes enables even less accurate controllers that miss an area to achieve high fitness through a process of feedback.  This is increasingly utilized because more time is available to perform these behaviors with increased digging accuracy. The net result is further improvement in fitness.

\subsection{Neuronal Analysis}

Several different methods have been used to analyze ANT solutions to identify emergent behaviors that solve the excavation task.  First we analyze neuronal and behavioral activity of a evolved tissue solution with a fitness of 0.99. It is difficult to discern what the controllers are doing based solely on output behaviors unlike previous task such as sign following and resource gathering \cite{Thangav2008,Thangav2010}. Hence we have used sensor-behavior combinations (Table~\ref{tbl:detect}) to detect instances of complex behaviors that have evolved to solve the task.  Further we identify the location and activity of these behaviors in the tissue topology.

\begin{figure*} [h]
    \centering
    \includegraphics[width=5.25in]{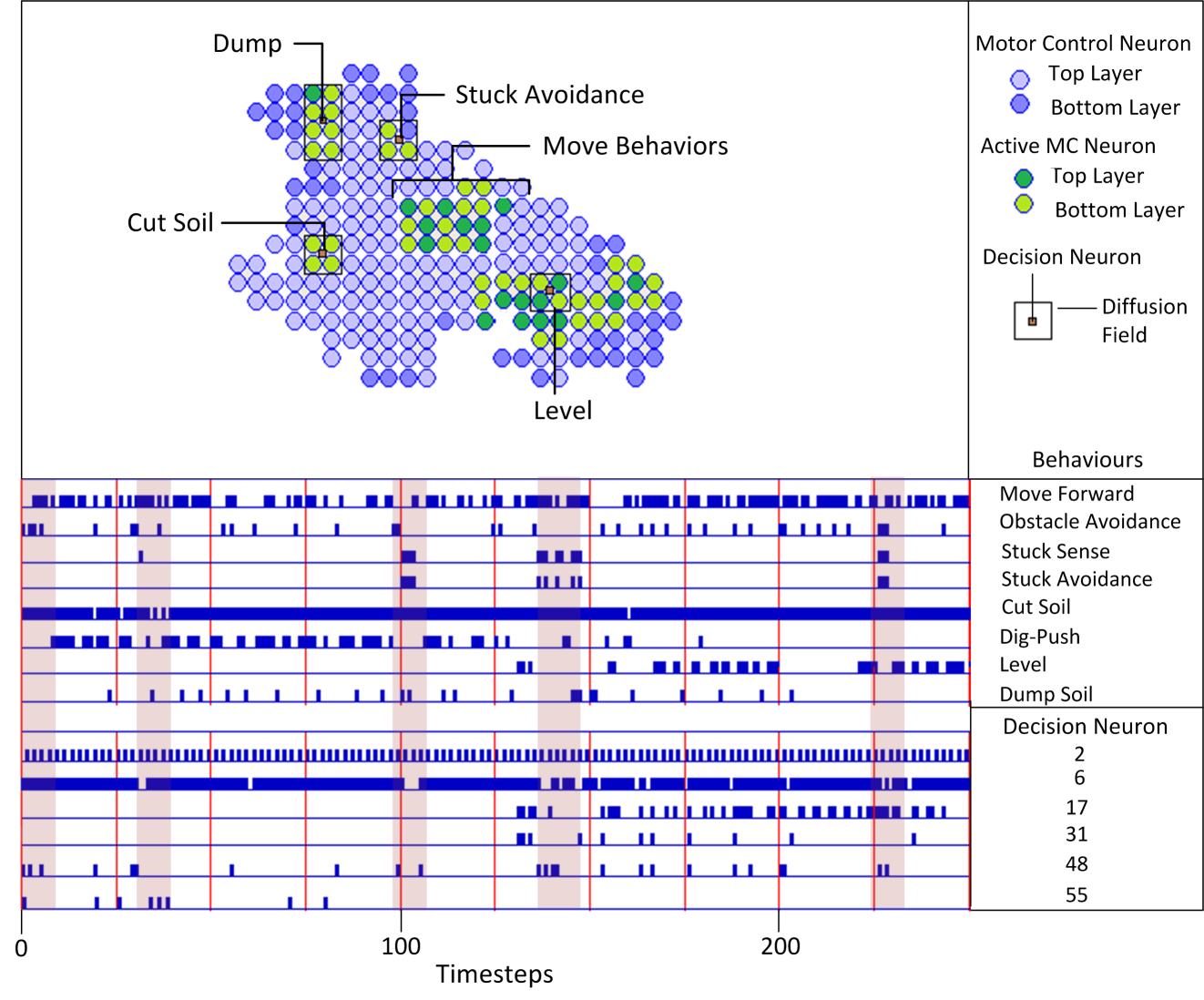}
 %\vspace{-12pt}
\caption{ANT topology and neuronal activity identifying important behaviors evolved for the excavation task.  Shaded columns highlight correlation in neuronal and behavioral activity during a typical simulation run. The controller was evolved using 4 robots in a $8 \times 8$ excavation area and has a fitness of 0.99. } \label{fig:behave_analysis}
 %\vspace{-5pt}
\end{figure*}

Figure~\ref{fig:behave_analysis} shows an ANT tissue with the location and activity of major behavioral centers.  As with other tasks evolved using ANT \cite{Thangav2008,Thangav2010}, the behaviors evolved as neurons modules are distributed within the tissue.  Interestingly, most decision and motor neurons remain dormant and don't encode for any identified behaviors.  This further suggests that these neurons remain neutral.  This is beneficial in evolution, where the right combination of mutations can first encode a function neutrally followed by a trigger mutation that unlocks this behavior.  The neutral neurons act as scaffolding for this transition.   With so many neurons remaining dormant, this permits the process to occur in parallel throughout the tissue.  Once a new behavior is turned on in a trial and error, explorative process, there is no guarantee that this behavior will provide a fitness advantage.  If it doesn't, this individual has reduced chance of survival through selection. However if this change results in a behavioral innovation that provides a significant fitness advantage, then the progeny multiplies and may then dominate the evolving population. This repeated process of trial and error continues, with innovations being accumulated much like building blocks until a high fitness is reached \cite{Thangav2010}.

These identified neurons modules have acquired specialized traits that are use to perform incision cuts (cut dig), scout, level, dump, obstacle avoidance and perform stuck avoidance behaviors (Figure ~\ref{fig:behave_analysis}). Analysis of decision neuron activity with sensor input and behavior output combination show  correlations. This suggests the behaviors are distributed among one or more decision and motor neuron modules.  These modules work cooperatively to encode tissue behavior.  This has a significant advantage.  For one, if a module is perturbed due to deleterious mutations, then another module that partly encodes for the behavior can recover some of this functionality.  Further, if multiple modules encode for the exact same functionality, then deleterious mutations have minimal effect.

At the bottom of Figure~\ref{fig:behave_analysis}, key events in a typical run lasting 250 timesteps are shaded. Observing the move forward behavior as with the other 12 behaviors from Table~\ref{tb:excavate_behaviors}, it is difficult to ascertain what is occurring due to the complex interactions of multiple robots and their behaviors.  Instead we obtain statistics of sensor-behavior output combinations from Table~\ref{tbl:detect} to better understand the solutions.

Several correlations are identified here, including between
stuck avoidance and decision neuron 6, leveling and decision neuron 31, obstacle avoidance and decision neuron 48, cut-dig and decision neuron 55.  Interestingly, decision neuron 2 behaves like a clock oscillating between its two binary states.  This oscillator behavior has been observed in other ANT solutions for different tasks \cite{Thangav2008,Thangav2010}. It is hypothesized that this is used to synchronize the various behavior modules distributed within the tissue.

\subsection{Behavioral Analysis}

\begin{figure*} [h]
    \centering
    \includegraphics[width=6.25in]{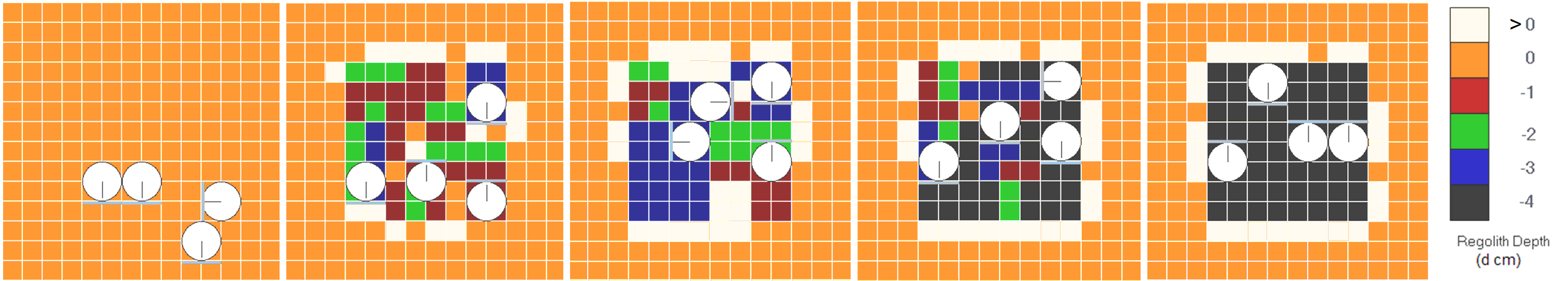}
 %\vspace{-12pt}
\caption{Simulation snapshots of an excavation task simulation (4 robots) after 0,50,75,100,170 timesteps.} \label{fig:ex_digging_screen}
 %\vspace{-5pt}
\end{figure*}

In a second approach, we analyze evolved solutions by observing robot controller behavior in the simulation environment (Figure \ref{fig:ex_digging_screen}).  The robots  scan the area in front at each timestep to determine whether its above or below the goal depth.  Once it is found to be above the goal depth it starts digging.  Because the robots are randomly positioned at the start, the excavated areas appear randomly throughout the work site.  The robot start digging by lowering their blades one level below ground and move forward one grid square forward.   Immediately afterwards, the controllers raise the blade to level and continue pushing soil at a level height until reaching a dumping area.  Once the robots detects a dumping area in front it pushes the soil forward, followed by a reverse and turn.  This results in the accumulation of piles of soil (berms) at the dumping area.

The controllers learn to perform layered digging (Figure~\ref{fig:ex_digging_screen}). The controllers do not lower the blade immediately to reach the goal depth, instead they perform an incision cut and then push soil layer by layer until the target depth is reached. As one robot removes one layer, another cooperates and pickups where the last one left off.  This process continues until the goal depth is reached. Multiple robots also cooperate by detecting one another through obstacle sensing and avoid getting in the way. Once another robot is sensed in front, the robot makes a turn.  With sufficient number of these interaction, many of the robots end up being in parallel tracks which minimizes obstructions from other robots. The dug areas that appear randomly scattered through the worksite at the beginning merge into one large area at the end.

The process of self-organization occurs with individual robots controllers sensing and manipulating the environment using local information.  The controllers can only sense local information. Although a global goal map may be present, each controller can only sense a localized region of the map at a time.  The initial appearance of excavated areas randomly throughout the work site, followed by their merger into one excavated area that meets the goal depth specifications is an example of the self organization.  This self-organization occurs due to cooperative behavior of multiple robots using local sensor input and performing only local actions.  The work done by one robot is not undone by another.

Analysis of the solutions suggests that the
controllers exploit templates by learning to correctly interpret dumping zones, ''don't care'' regions, depth
relative to the specified excavation blueprint.  These controllers perform interpretation by taking the discernable sensor input and determining the proper action to perform, such as move forward or backward and  whether to lower, level or raise the blade (Figure \ref{fig:ex_digging_screen}). With the goal map, if the goal depth is lower than the current soil depth, then the interpreted action is to dig by lowering the blade below ground and pushing.

Communication between robots
occurs through manipulation of the environment in the form of
stigmergy.  Manipulation of the environment occurs by
excavating a region or dumping excavated material. Each robot interprets the blueprint and determines whether to deeper, back-fill or move to the next location.  However these robots can't typically dig to the goal depth all at once and hence have to dig material layer by layer.  By this robots can only remove part of the soil at a time before having to dump at the dumping locations.  It is up to another robot to determine that the goal depth has not been reached and pickup where the last robot left off, which is a form of implicit communication mediated through the environment.  Furthermore once the goal depth is reached at a grid square, any other robot that comes and sense the grid square interprets that the goal depth is reached.

Through this implicit communications, cooperative actions also occur.  That is the excavation work done by one robot is not undone by another.  Initially small digging areas form, but in time through cooperative actions these digging areas merge into one region that matches the goal depth.  In addition, the ANT controllers exploit the ability to sense the depth of soil relative to wheel depth (Figure~\ref{fig:ex_digging_screen}).  This enables each robot controller to sense whether it is excavating deeper or backfilling at the current
depth.  The ability to backfill while useful,
can also undo the effort of other robots excavating at different
depths.  Sensing and avoiding this scenario is a form of cooperation.

The robots also have the ability to sense the relative position of a nearby robot much like
radar.  This feature is exploited to
to avoid collisions confirmed from Figure~\ref{fig:task_evo}b, when a series of output behaviors
such as `move forward' and `turn left' is applied in sequence.
Although obstacles can be detected using front sonars, there exists blind spots to the extreme right and left
making it difficult to detect and react to obstacles when a
sequences of behaviors are executed.

%\begin{figure}
%    \centering
%    \includegraphics[width=6.0in]{figures/combine}
%\caption{Scaling of ANT based solutions for varying depth (a) and 4
%robot solution for varying excavation area (b).}
%\label{fig:ex_combine}on
%\end{figure}
\subsection{Summary on How the Controller Works}

\begin{figure*} [h]
    \centering
    \includegraphics[width=6.25in]{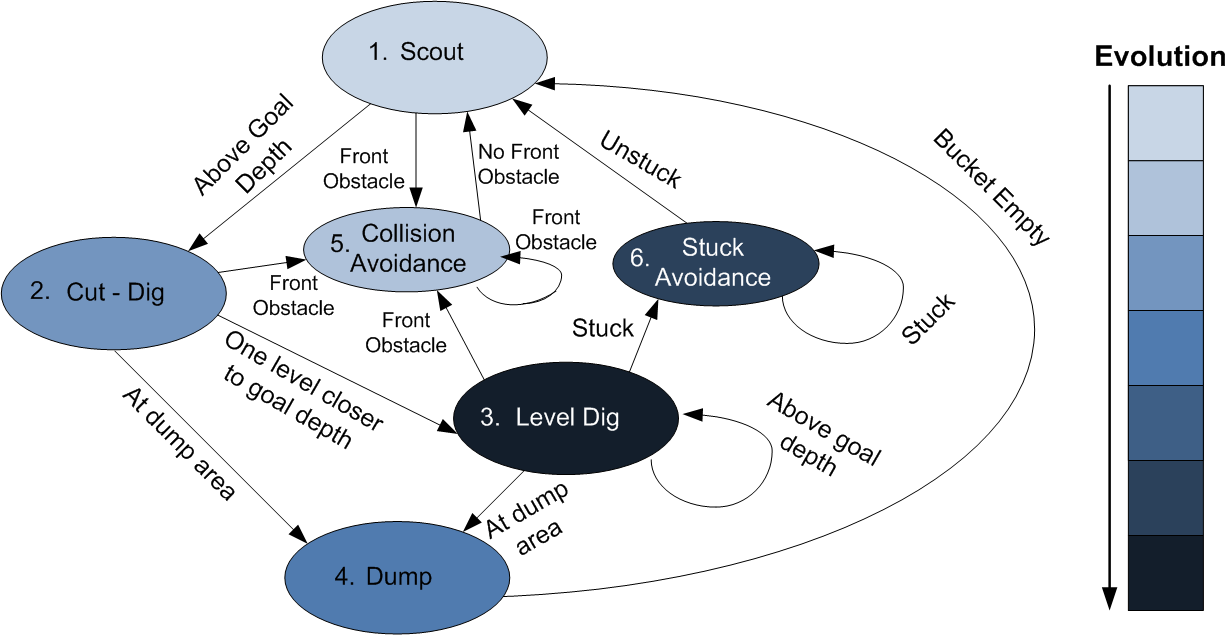}
 %\vspace{-12pt}
\caption{Representation of Major Behavior States Evolved by the ANT Solution Controllers.  Shading indicates the order in which these complex behaviours are evolved.} \label{fig:works}
 %\vspace{-5pt}
\end{figure*}

We used different techniques to analyze how complex behaviours emerge within the ANT controllers using neuronal analysis of controller that attains a fitness of 0.99.  Within this tissue, we identify portions of the tissue that are active and find different specialized modules emerge to perform the required subtasks required to perform the overall task.  We are also able to trace the evolution of these behaviours to a series of critical events, when the controllers discover new capabilities leading to substantial improvement in fitness performance.  Here we summarize our finding in Figure~\ref{fig:works} in the form of a finite state machine.

The controller during typical operation (1) scout (moves around) to find regions that are above the goal depth.  Once found the blade is lowered to (2) cut/dig into the soil, moving forward and then transitioning into (3) ``level digging.''  As the bucket is filled or if the robot nears a dumping area, the soil is dumped (4).  Along the way, robots perform collision avoidance (5).  This in fact appears to be a series of behaviours that minimizes robot to robot interaction, enabling the robots to work in parallel and to maximize area coverage. This ability to minimize direct robot to robot interactions influences the scouting behaviour state and decision of where to perform cut-dig. Along the way the robot may also need to avoid getting stuck (6) when pushing too much regolith.  As will be shown later, they take a non-greedy approach to excavation, where it is more efficient for individuals to temporarily give up when stuck (by unloading the pile of regolith being pushed) and resume later.

We also summarize how these behaviours have evolved.  The ability to scout and perform basic collision avoidance is evolved first.  This is followed by incision cuts.  Initially the incision cuts are performed randomly through trial and error, in addition to dumping the regolith.  In turn controllers evolve to interpret the signs correctly to dump at the designated locations.  However this process is still inefficient, limiting the amount of regolith being excavated or causing to dig below the designated goal depth.  With controllers evolving to excavate more regolith, they can get stuck pushing too much.  The controllers in turn evolve stuck avoidance behaviours used to abort excavation when stuck and start over from a different direction.  A critical step is reached when the controllers ``discover'' the level dig behaviour that allows pushing of regolith over longer distances and enables excavation and transport of significantly more regolith to dumping area resulting in a significant leap in fitness.

\subsection{Scalability}

First we consider performance of an intuitive hand-coded robot controller (see Appendix 8.1 for program description) for varying number of robots.  The hand-coded controller shows the best performance for a single robot scenarios, but steadily decreases in performance for increasing number of robots.  For this controller it is assumed that robot need to avoid one another and hence obstacle avoidance behaviors are included.  The results show this assumption is not optimal and results in poor performance for increased number of robots.  The controller is not effective in exploiting parallelism.  This suggests, cooperative behaviors are necessary to exploit increased number of robots, and do not just involve intuitive obstacle avoidance procedures.

\begin{figure} [h]
    \centering
    \includegraphics[width=4.5in]{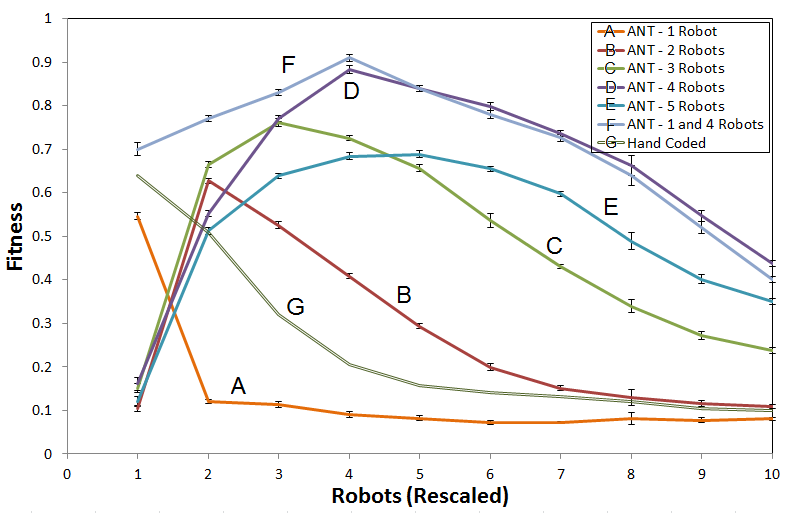}
% \vspace{-12pt}
\caption{Excavation performance of ANT based solutions evolved using 1 to 5 robots $(8
\times 8$ excavation area) applied to scenarios using 1 to 10 robots.}
\label{fig:ex_scaleup}
 \vspace{-5pt}
\end{figure}

The fittest ANT solutions from the simulation runs shown in
Figure~\ref{fig:ex_evo} are applied to this new setting, by varying the number of robots
while holding the digging area constant (Figure~\ref{fig:ex_scaleup}). Taking the controller evolved for a single robot and running it on a multirobot system shows a steep
degradation in performance.  This is expected since the single-robot
controller lacks the cooperative behavior necessary to function well
within a multirobot setting.  For example, these
controllers fail to develop `collision avoidance' behaviors.
Similarly, a multirobot system scaled down to a single robot setting
also shows a degradation in system performance.  With the multirobot
system, controllers have evolved to exploit and depend on
cooperative actions to complete the task; thus when the environment
is abruptly changed the controllers perform poorly.  With more than 4 robots for a $8 \times  8$ area, the problem of antagonism arises when multiple robots trying to perform the same task interfere with one another and reduce the overall
efficiency of the group.  The key here is number of robots selected during training.  Proper selection of robots better enables the controllers to be scalable.  Further we extend the comparison by mixing number of robots scenarios during training.  For example mixing training scenarios of 1 and 4 robots shows better scalability performance than one and four robot solutions.  In addition this approach shows better performance than the intuitive hand-coded solution when using a single robot.

\begin{figure} [h]
    \centering
    \includegraphics[width=4.5in]{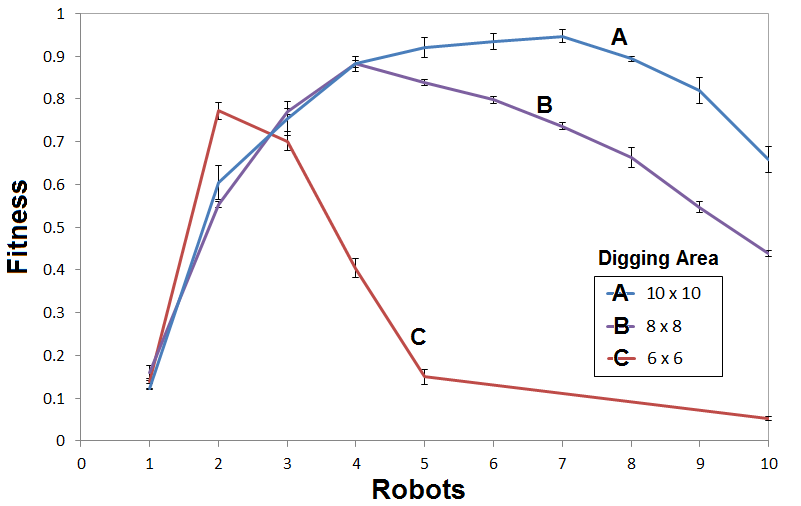}
% \vspace{-12pt}
\caption{Excavation performance ANT based solutions evolved using 4 $(8
\times 8$ excavation area) applied to resized excavation area and number of robots.}
\label{fig:antagonism_confirm}
\vspace{-5pt}
\end{figure}

This effect of antagonism on performance is further supported in Figure~\ref{fig:antagonism_confirm} by varying excavation area and applying an ANT solution evolved using 4 robots on a $8 \times 8$ area to the $6 \times 6$ and  $10 \times 10$ digging area.  For the $6 \times 6$ area, peak performance is achieved using only 2 robots, but show a larger drop in performance for increased number of robots, showing the increased effect of antagonism for a smaller area.  In addition, increasing the area results in the peak performance reached with 7 robots and performance drops in smaller increments.  These two experiments confirm the problem of antagonism with these multirobot systems.  Furthermore, the ANT solution show higher fitness than during training for the large digging area with increased number of robots

\begin{figure} [h]
    \centering
    \includegraphics[width=3.25in]{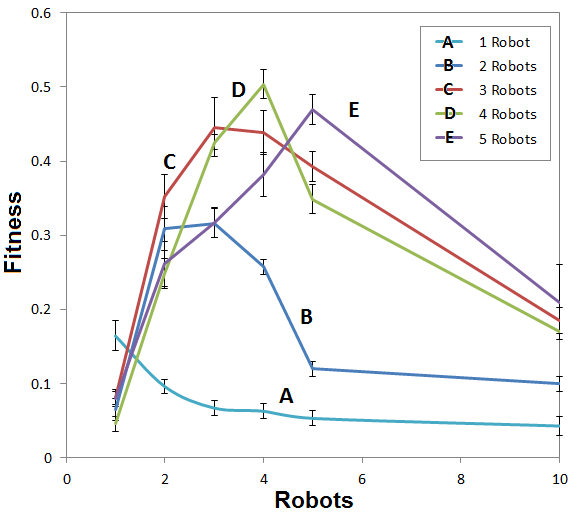}
 %\vspace{-12pt}
\caption{Excavation performance of the fittest standard neural networks evolved using between 1 and 5 robots applied to scenarios using 1 to 10 robots.} \label{fig:rescaled_std}
\vspace{-5pt}
\end{figure}

As noted earlier, a key factor in developing adaptive controllers is scalability.  Scalability is critical, because in real world scenarios, one or more robots may be disabled or unavailable and the approach needs to effectively complete the task. Taking the evolved standard neural network controllers and applying them on $n$ robots is shown in Figure~\ref{fig:rescaled_std}.  Comparing the results with Figure~\ref{fig:ex_scaleup}, standard neural networks controllers show poor scalability performance. In comparison, ANT shows a two-folds performance advantage.  Controllers evolved for 4 robots, shows highest fitness for a 4-robot scenario.  In addition, controllers evolved for a single robot show poor scalability applied on increased number of robots.  Alternately, controllers evolved with  4 robots show relatively better performance on average than other training scenarios.  This indicates that in a real world scenario, ANT controllers are more robust to handle uncertainties in the number of active robots towards completing a task.

%[Varying Excavation Depth] [Varying Excavation Area]
%\begin{figure} [h]
%  \centering
%  \subfigure{
%    %\label{labelname 1}
%    \includegraphics[height=2.5in,width=3.3in]{figures/fig_comb1}
%  }
%  \subfigure{
%    %\label{labelname 2}
%    \includegraphics[height=2.5in,width=3.0in]{figures/fig_comb2}
%  }
%  \vspace{-12pt}
% \caption{Scaling of ANT based solutions for varying depth (left) and 4
%robot solution for varying excavation area (right).}
%  \label{fig:ex_combine}
%   \vspace{-5pt}
%\end{figure}

\begin{figure} [H]
    \centering
    \includegraphics[width=4.0in]{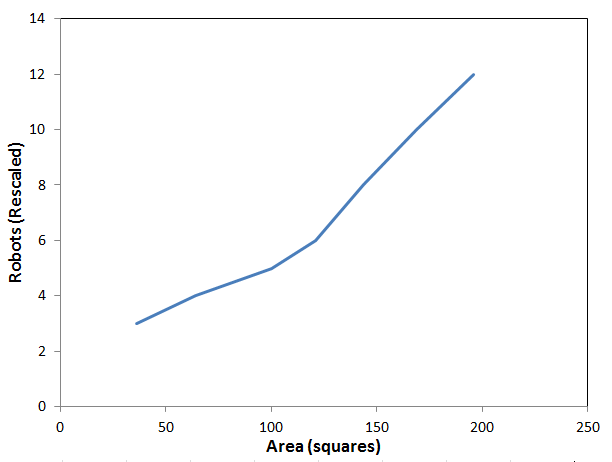}
%     \vspace{-12pt}
\caption{Excavation performance of ANT solutions evolved using 4 robots on a $8
\times 8$ excavation area for varying excavation area.} \label{fig:ex_scale}
% \vspace{-5pt}
\end{figure}

\begin{figure} [H]
    \centering
    \includegraphics[width=4.0in]{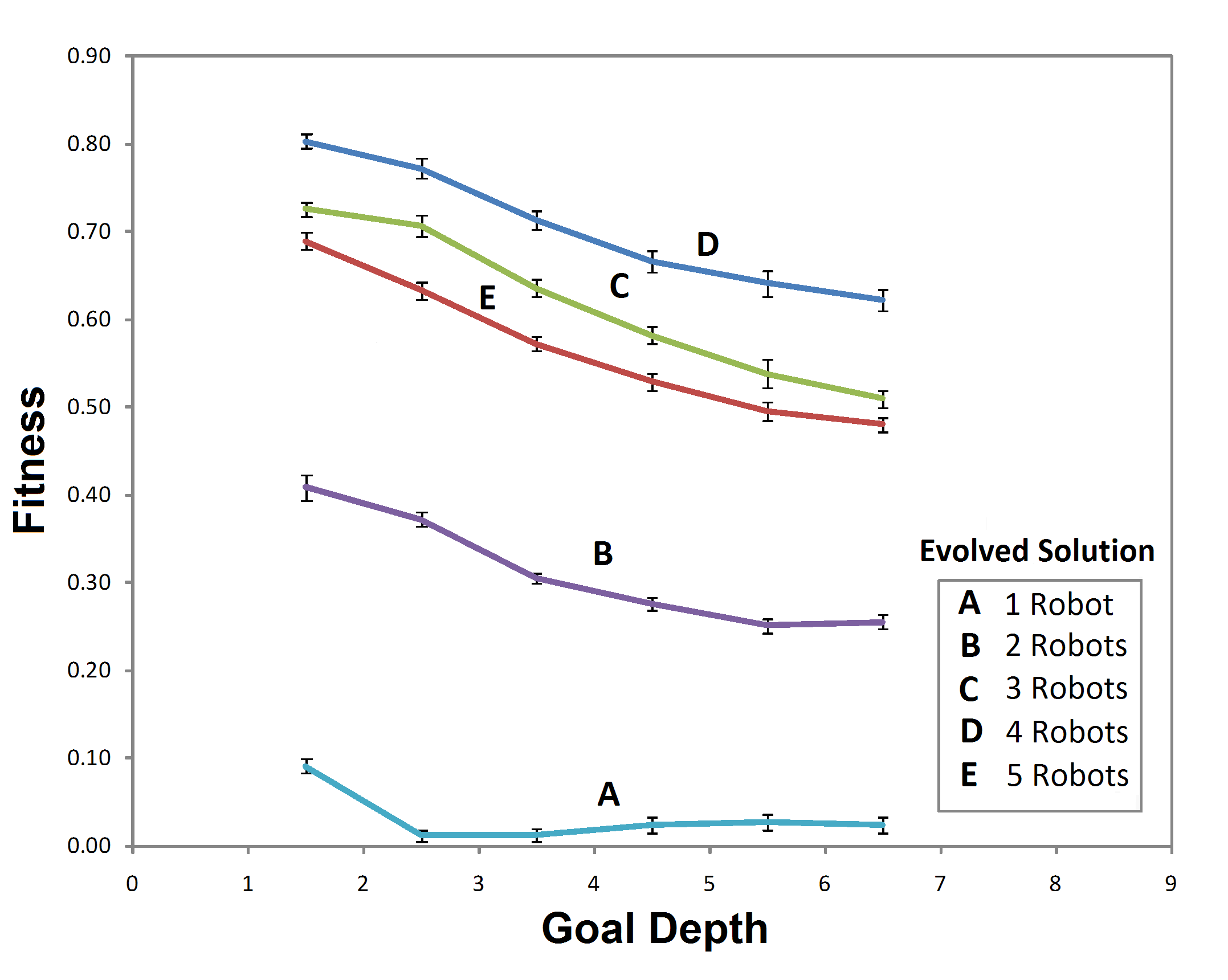}
%     \vspace{-12pt}
\caption{Excavation performance of ANT based solutions for varying depth taken after 10,000 Timesteps} \label{fig:ex_depth}
% \vspace{-5pt}
\end{figure}

Focusing on the ANT controllers, it is interesting to note that the controllers trained with 4 robots
for an $8 \times  8$ digging area perform considerably better on average
than solutions trained for other number of robots.  It is
evident that this solution perform better for increased goal depth
(Figure~\ref{fig:ex_depth}) and show better performance
than other solutions for increased excavation area. The optimal
ratio of robots to digging area using solution trained
with 4 robots is shown in Figure~\ref{fig:ex_scale}.

These simulation experiments suggest that there exists an optimal set of training
conditions that enables controllers to evolve improved scalability.  Although
the controllers may be better adapted to antagonism under higher
training densities with improved obstacle avoidance techniques,
these behaviors may not be well tuned to completing the overall
objectives effectively.  This optimal condition is dependent on task duration. Furthermore, the optimal density is beneficial when the task is time limited.  Given enough time, the suboptimal solution can attain the same fitness but consumes more energy.

\subsection{Selection for Multirobot Behavior}

\begin{figure} [h]
    \centering
    \includegraphics[width=4in]{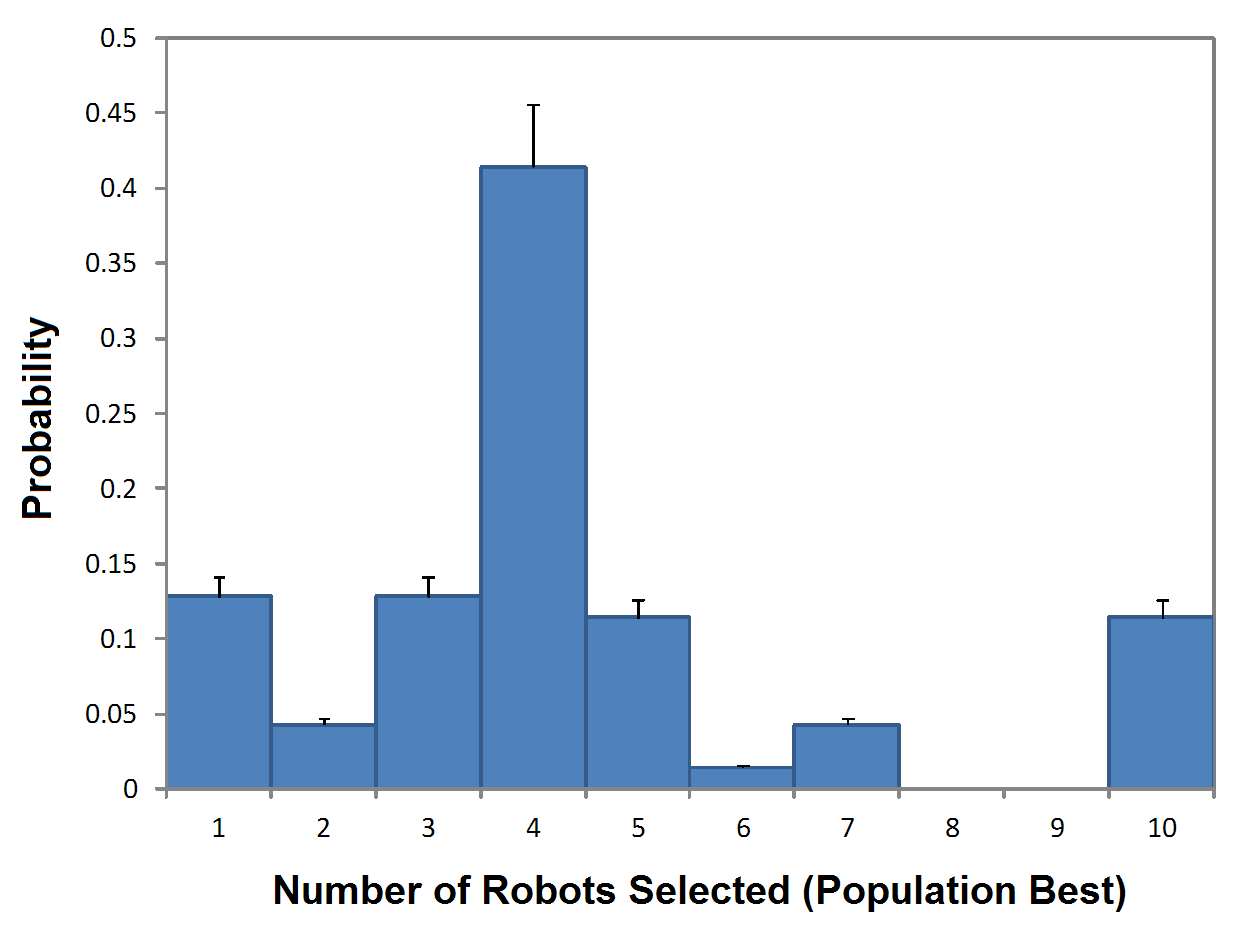}
  %   \vspace{-12pt}
\caption{Histogram of number of robots selected, where $N$ the number of robots is evolved as parameter (right).} \label{fig:ex_sel}
 %\vspace{-5pt}
\end{figure}

Based on Figures~\ref{fig:ex_scaleup} and~\ref{fig:ex_scale}, the ANT controllers trained for a 4-robot solution produces solutions that show better rescalability than other initial conditions for a $8 \times 8$ excavation area. Can evolutionary techniques be used to determine both a multirobot controller and optimal system parameters (such as number of robots)
simultaneously? For this experiment, the number of robots, $N$, is introduced as an evolvable variable within ANT. This is akin to a population reproduction rate, where certain species have more children at a time than others. A histogram in Figure~\ref{fig:ex_sel} shows the distribution of $N$ among the population best averaged over 30 runs. If there were no selection pressure on $N$, then the histogram should be a uniform distribution.  The results suggests evolutionary selection tends towards the observed optimal number of robots (Figure~\ref{fig:ex_scaleup}).

The question then is why is the system tending toward the optimal number of robots per given area?  It is not obvious that controllers gravitate towards an optimal setting and stay there.  For one, if the optimal solution is sensitive to damaging mutations, then the population may gravitate to a more stable suboptimal solution.  However, solutions evolved under the optimized condition rescale better.  Both traits provide an advantage, helping to gravitate solutions toward this setting and remaining there.  An individual having adapted under this optimal setting is better adept at handling a deleterious mutation of $N$ robots that could either result in increased or decreased number of robots.

This technique of evolving a controller and other system parameters concurrently reduces the need to analyze the rescalability performance and antagonism. Because the system tends towards the optimal number of robots, the problem of antagonism is limited. However owing to the stochastic nature of the search process, the optimal number of robots can only be obtained with statistical certainty after repeated evolutionary runs.

\section{Proof-of-Concept Experiments}

\subsection{High Fidelity Lunar Simulator}

This section describes work on porting ANT controllers onto excavation robots in the Digital Spaces$^{\text{TM}}$ simulator.
Digital Spaces$^{\text{TM}}$ is an off-the-shelf, commercial, high-fidelity 3-D simulator that simulates the low gravity lunar surface. The Balovnev Soil Interaction Model~\cite{balovnev,nader} is used to simulate the deformable terrain. Deformable terrain modeling allows for accurate regolith-tool and wheel-terrain
interactions. This virtual approach facilitates prototyping and testing of alternative digging concepts and can potentially reduce hardware experiment costs especially of off-world environments.

The simulated robots are equipped with a front loader, are holonomic and deliver 300 W average power. The robot model shown in Section~\ref{sec:robot_model} is ported to the Digital Space environment.  Details of the approach are found in \cite{nader}.  The $(x,y)$ position, depth ($z$), tilt ($R_1$) are built in variables.  Using a discretized mesh of the surface terrain and $(x,y,z)$ of the robots, $Z_1$\ldots$Z_4$ and $E_1$\ldots$E_2$ are computed.  Relative position of the robots are used to compute the state of the front obstacle avoidance sensors ($S_1$), heading of nearest robot $H_1$ and distance to nearest robot, $D_1$.  The blade positions are simulated on the front loader by defining four set positions to match the bull dozer blade settings, including above, level, below ground and home position. The blade load, $L_1$ is computed using the soil interaction model~\cite{balovnev,nader}.

\begin{figure} [h]
    \centering
    \includegraphics[width=6.25in]{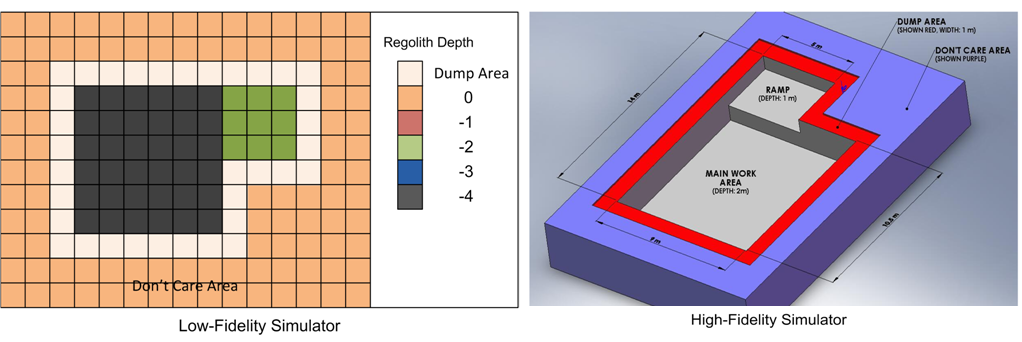}
 %\vspace{-12pt}
\caption{Excavation blueprint of a landing pad.  Low fidelity training simulator (top). High fidelity Digital Spaces$^{\text{TM}}$ simulator (bottom).} \label{fig:digi_goalmap}
 %\vspace{-5pt}
\end{figure}

\begin{figure} [H]
    \centering
    \includegraphics[width=5.0in]{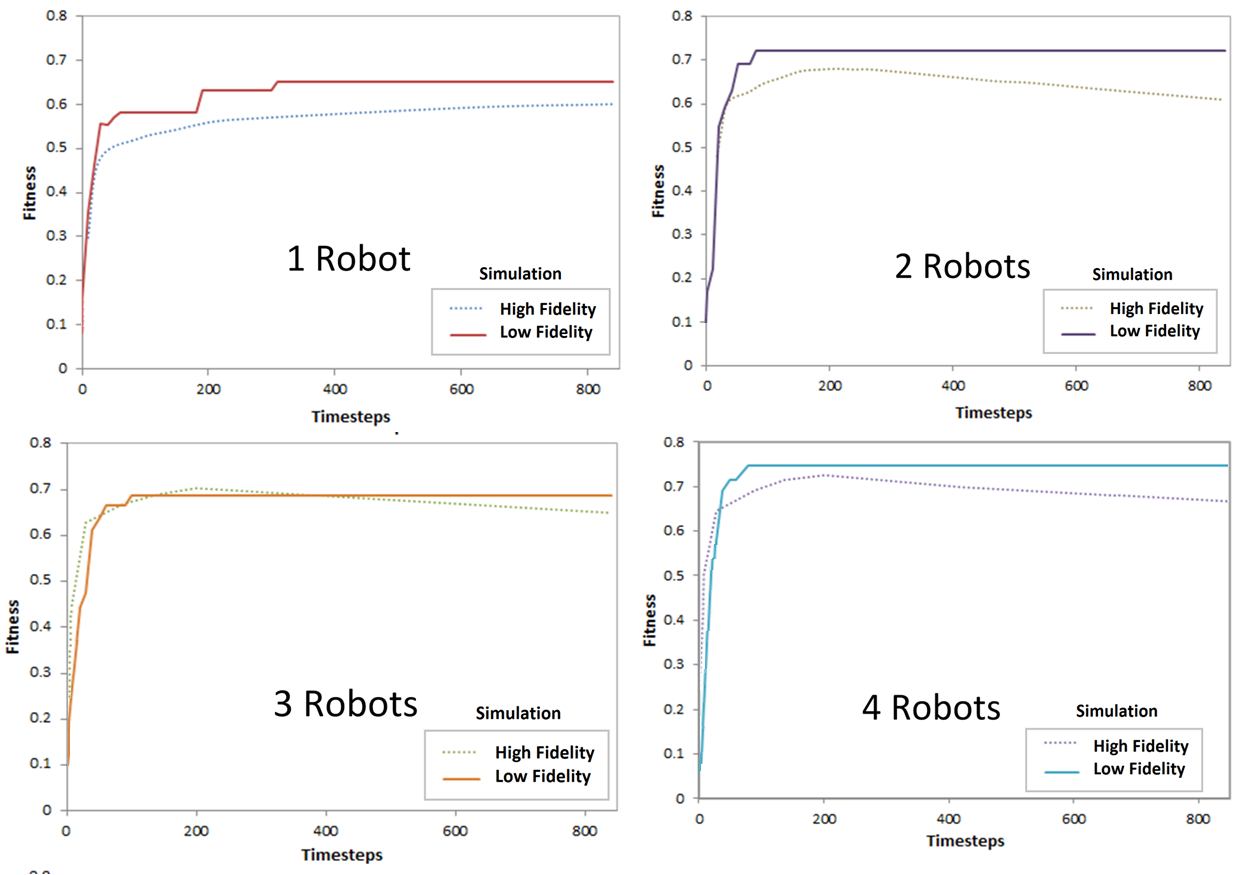}
 %\vspace{-12pt}
\caption{Fitness comparison of ANT controllers in the low-fidelity training simulator and high-fidelity Digital Spaces$^{\text{TM}}$ simulator. } \label{fig:digi_fitness}
 %\vspace{-5pt}
\end{figure}

\begin{figure} [H]
    \centering
    \includegraphics[width=6.2in]{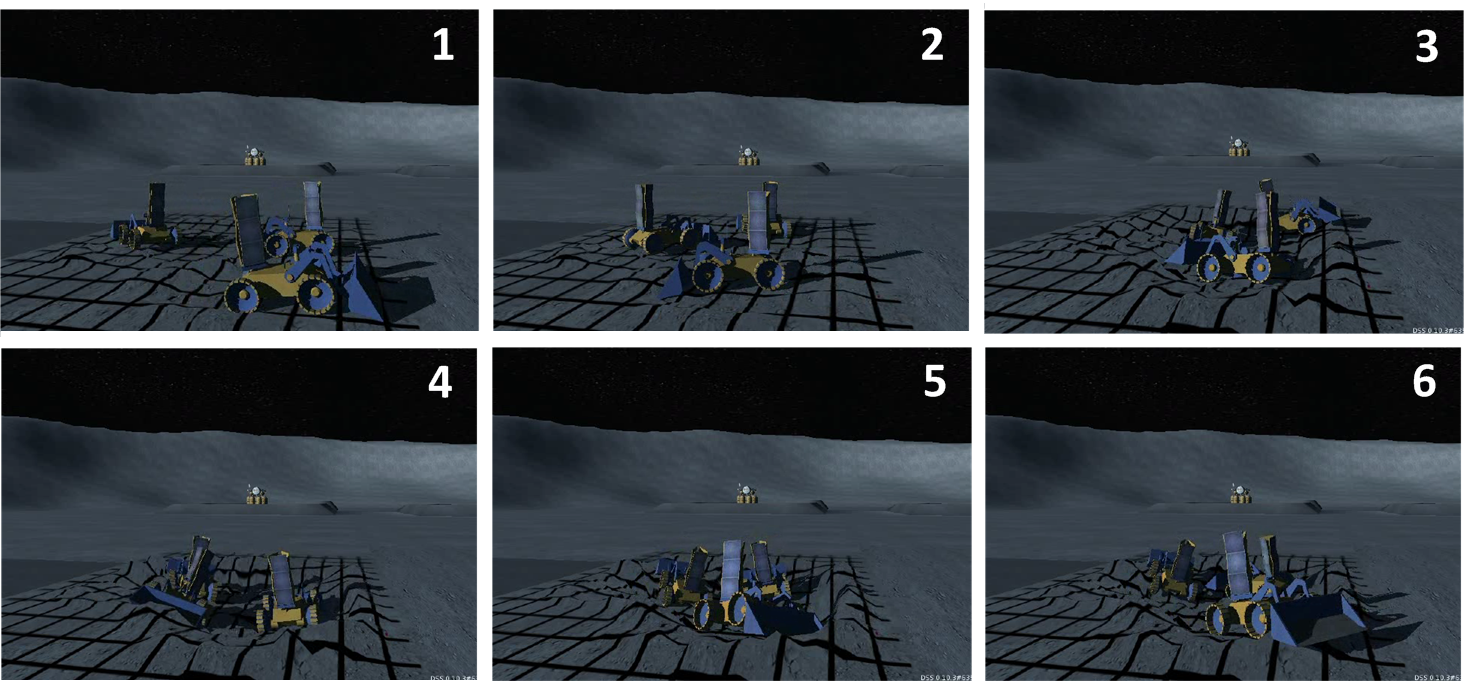}
  %   \vspace{-8pt}
\caption{Digital Spaces$^{\text{TM}}$ simulation of three robots using a four-robot ANT controller solution to perform excavation on simulated lunar terrain.  The robots unlike under training conditions use a front-loader bucket instead of a two-way bulldozer blade.  The excavation blue print includes a hole and a ramp for the robot to enter/exit surrounded by the dumping area.} \label{fig:ex_nader}
% \vspace{-5pt}
\end{figure}

To ensure consistency between the training simulations and the Digital Spaces virtual world, fitness is monitored for 1, 2, 3, and 4 robots applied to an excavation blueprint of  a landing pad shown in Figure~\ref{fig:digi_goalmap}. The results are shown in Figure~\ref{fig:digi_fitness} right. The excavation blueprint consists of a landing pad $9 \times 10.5 \times 2.0$ m deep connected by a $5.0 \times 3.5 \times 1.0$ m deep region for a ramp to exit/enter the landing pad. Comparison with the training simulator (Figure~\ref{fig:digi_fitness}) shows that the results correlate well. After reaching a peak fitness at about 200 timesteps for all except the single robot scenario, the fitness drops gradually in the Digital Spaces simulations. The training simulator results (Figure~\ref{fig:digi_fitness}) level off within 200 timesteps (except for the single robot scenario). In Digital Spaces simulations, the fitness then gradually drops after the final goal depth is reached. In this case, the robots continue to move around the work area with blade height set to level. Theoretically, this means the robots do not dig any regolith, but in reality, they skim off small amounts of material, going slightly below the goal depth over time.  Video snapshots of a typical simulation is shown in Figure~\ref{fig:ex_nader}).

%\begin{figure*} [t!]
%    \centering
%    \includegraphics[width=6.75in]{double_graph3}
 %%\vspace{-12pt}
%\caption{(Left) Fitness Performance Comparison of ANT Based Solutions
%for between 1 and 5 Robots with varying time. (Right) Fitness performance vs. time of the evolved ANT %controllers using Digital Spaces$^{\text{TM}}$. } \label{fig:digi_fitness}
 %\vspace{10pt}
%\end{figure*}

\subsection{Controlled Field Experiments}

\begin{figure} [H]
    \centering
    \includegraphics[width=4in]{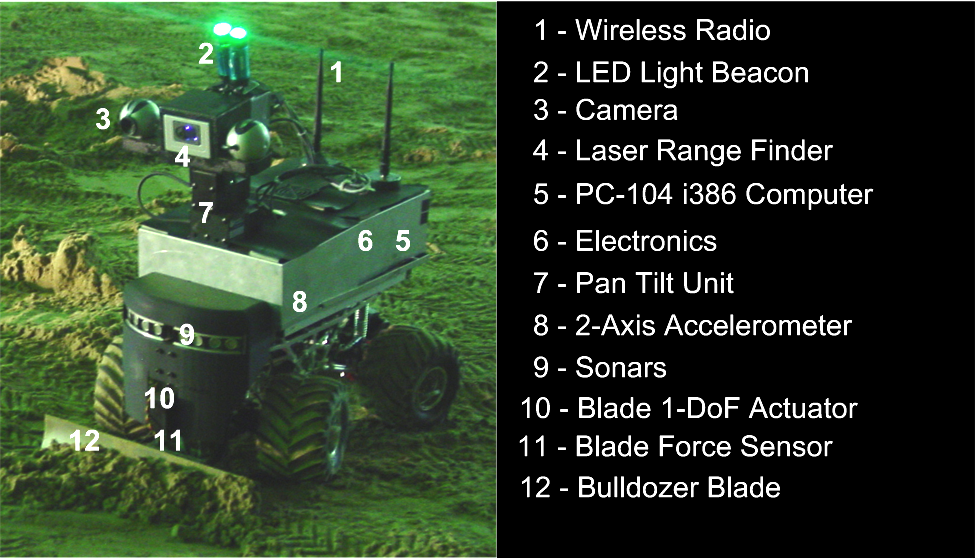}
%     \vspace{-12pt}
\caption{An Argo robot equipped with a 1 DoF bulldozer blade,
a laser range finder,a pair of Logitech$^{\circledR}$ Quickcams$^{\text{TM}}$ affixed to the pan-tilt unit and an assortments of other sensors. } \label{fig:argo_excavator}
% \vspace{-5pt}
\end{figure}

This section describes work performed to test the ANT excavation controllers on real robots.  The robots were tested inside a 50 m dome, containing loose sand.  The robots were tested under low lighting conditions representative of a lunar region shadowed due to cratering.  In addition the ambient temperature was between -5 $^{o}$C and -10 $^{o}$C.  The robot model shown in Section~\ref{sec:robot_model} is ported onto a team of UTIAS Argo class robots (Figure~\ref{fig:argo_excavator}).  The robots are approximately 15-kg, non-holonomic and are four wheel drive.  Each robot is equipped with a 1-DoF servo actuated bulldozer blade system, a PC-104 Intel 80386 computer, an assortment of sensors and actuators. The drive and actuation system is powered by two 8V, 4500 mAh lithium ion batteries.  The electronics and computers are powered by four 7.5V lithium ion batteries.

\begin{figure} [h]
    \centering
    \includegraphics[width=4in]{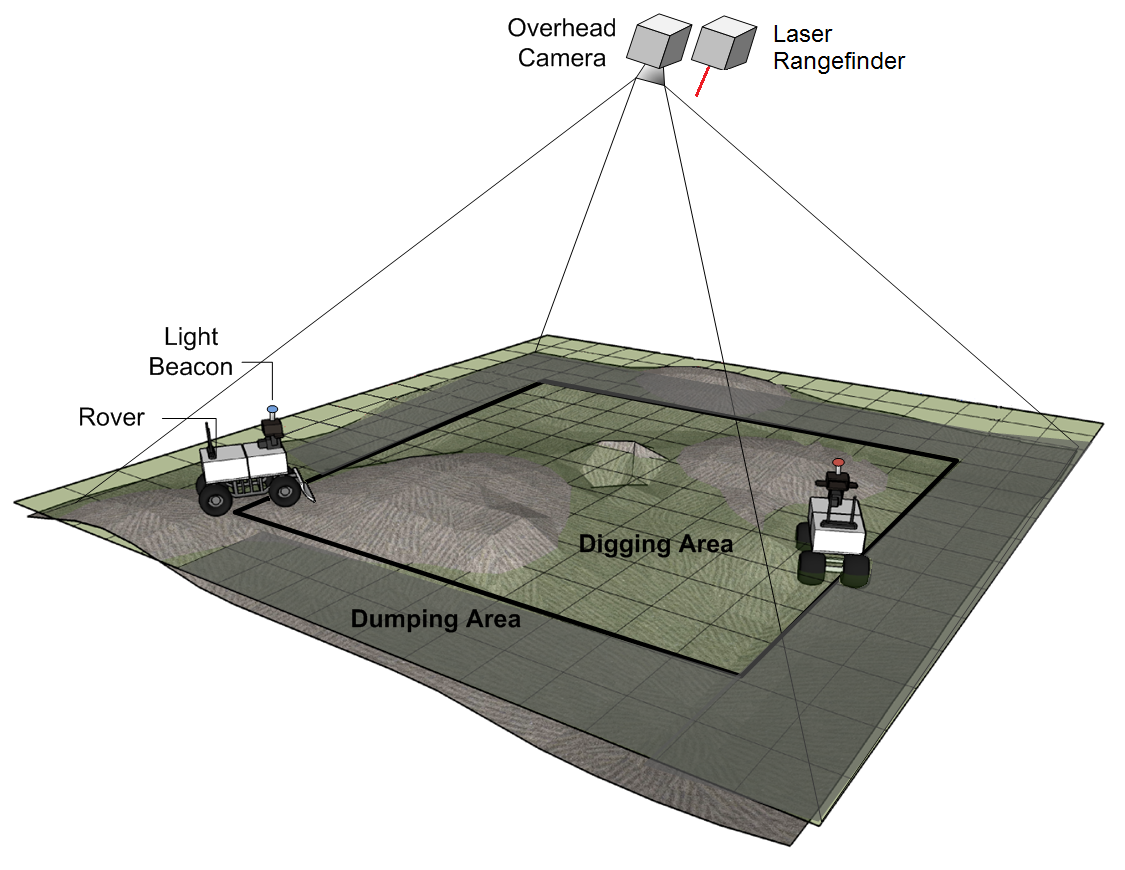}
%     \vspace{-8pt}
\caption{Layout of the experimental system, showing excavation robots mounted with light beacons localized using an overhead camera system.} \label{fig:overhead}
% \vspace{-5pt}
\end{figure}

An overhead camera is used to determine the $(x,y)$ location of the robots by tracking LED light beacons (Figure~\ref{fig:overhead}).   The variable $z$  is estimated using an overhead laser range finder that performs periodic scans of the worksite.  Using the estimated $z$ value, the robot equipped with a Leica laser range finder is used to compute $Z_1$\ldots$Z_4$ and $E_1$\ldots$E_2$. The onboard laser range finder and sonars are used to  detect obstacles at the front ($S_1$). Heading and distance of nearest robot, $H_1$ and $D_1$ respectively are computed using the overhead camera. $R_1$, the robot tilt is measured using an onboard 2-axis accelerometer.  A webcam  is used to determine $U_1$, i.e. whether the robot is stuck or not by simple comparison of consecutive images. On the robot, fine blade adjustments are made using PID control to ensure a constant force is met. The robot can handle a maximum of 10 N push force and this is rescaled to integer values between 0 and 4 for the blade load, $L_1$.  The excavation blueprint used in shown in Figure~\ref{fig:explain} (right).

\begin{figure} [H]
    \centering
    \includegraphics[width=6.2in]{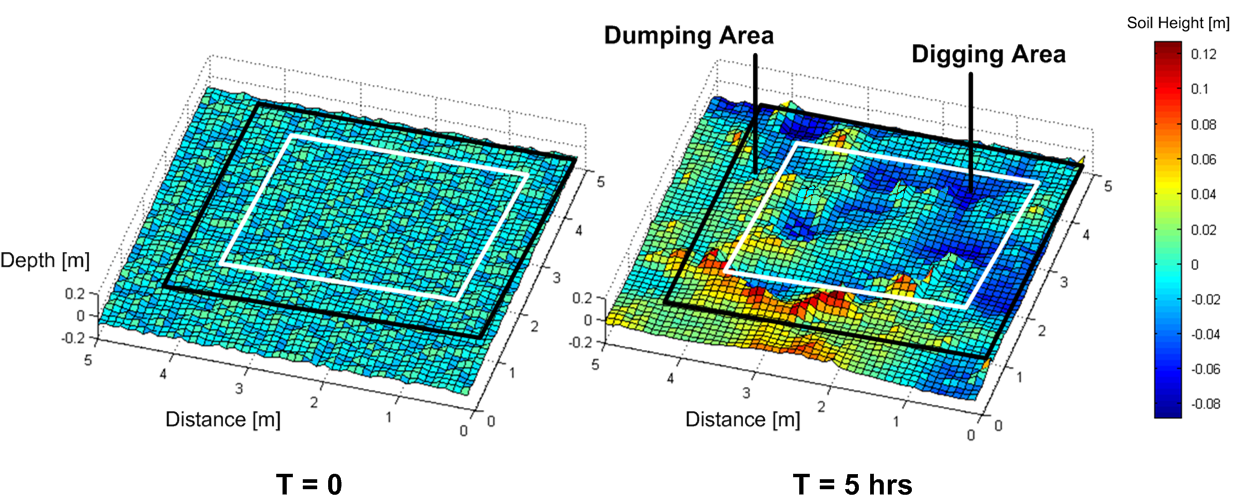}
 %    \vspace{-12pt}
\caption{An overhead laser scan showing the difference in soil height before and after the excavation experiment.} \label{fig:dig_progress}
 %\vspace{-5pt}
\end{figure}

\begin{figure} [H]
    \centering
    \includegraphics[width=6.2in]{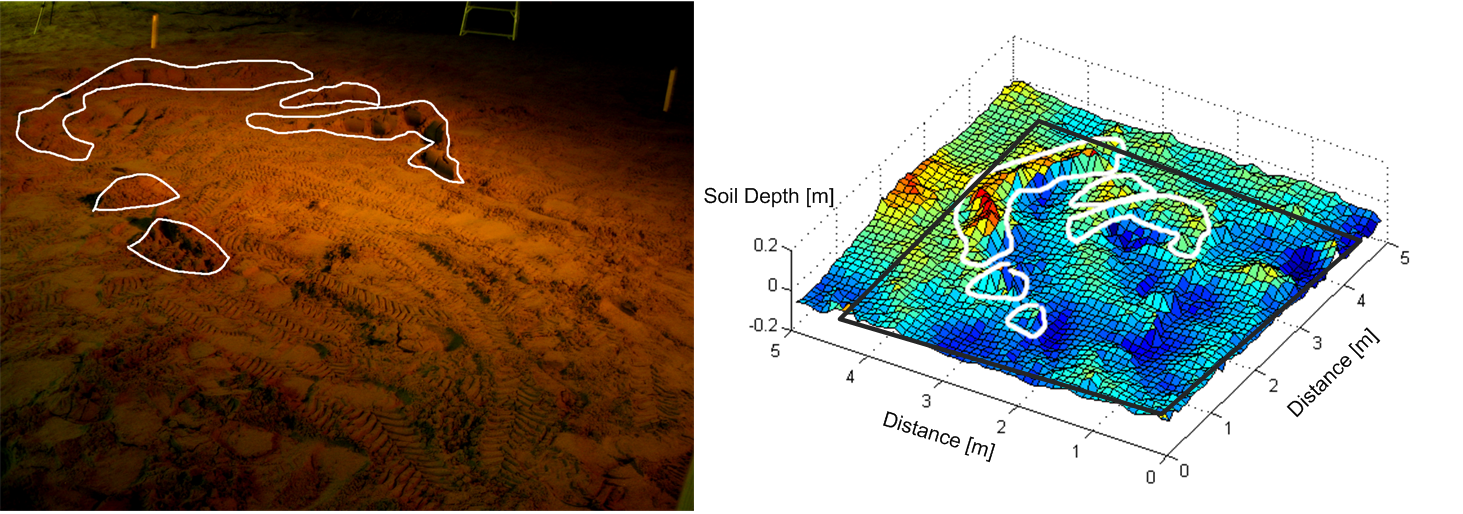}
    % \vspace{-8pt}
\caption{Photo of the work site after excavation (left).  Mounds of dirt (berms) highlighted. 3-D Laser scan of worksite with berms highlighted (right).} \label{fig:ground_photo}
 %\vspace{-5pt}
\end{figure}

Figure~\ref{fig:dig_progress} shows 3-D laser scans of the soil moved from a 5-hour excavation experiment with 2 robots.  Comparison of the laser scans show an estimated 0.7m$^3$ of soil volume was displaced in the test area.  This estimate is performed by averaging the regolith dug and dumped in the vicinity.  This measurement is typically an underestimate, because the regolith at the start of the experiment is uncompressed and after the experiment, with multiple rover traversal around the regolith becomes quite packed.  The robots start excavating holes randomly in the digging area.  These holes in turn merge to form one big pit.  In this excavation run, three sides of the pit wall have formed and are clearly visible from ground photographs and overhead 3-D laser scan (Figure~\ref{fig:ground_photo}).

Ground truth measurements show that the maximum depth reached is between the height of the berm and lowest point of the dig area to be 15 and 20 cm.  At the target depth, the regolith is quite packed.  This is despite the fact that the  robots are underpowered for the task resulting in the robots frequently getting stuck, backing out and attempt to restart digging, thus slowing the overall excavation process.  On most occasions the robots correctly interpret the excavation blueprints and dump dirt in the specified dumping areas as shown from the 3-D laser scans (Figure~\ref{fig:dig_progress}).

\begin{figure} [H]
    \centering
    \includegraphics[width=3.25in]{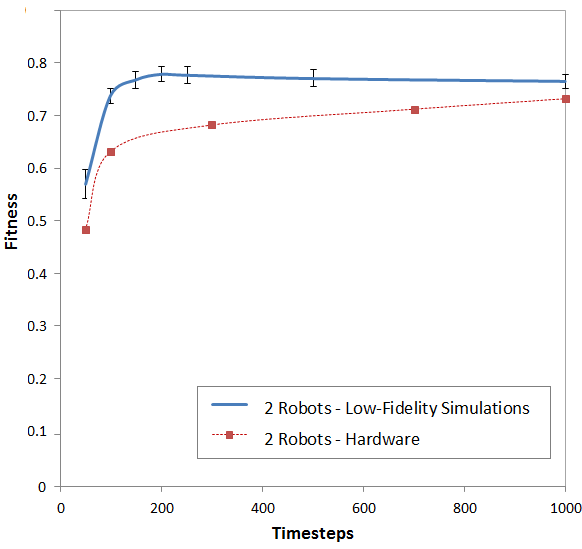}
 %\vspace{-12pt}
\caption{Fitness during a typical excavation run of an ANT controller evolved using 4 robots ($8 \times 8$ area) applied to 2 robots on the training simulator and real robots.} \label{fig:hardware_sim}
 %\vspace{-5pt}
\end{figure}

Fitness comparison between the low-fidelity training simulator and hardware experiments for the two-robot scenario is shown in Figure~\ref{fig:hardware_sim}.  In the training simulator, the steady state fitness is reached in shorter time than the hardware experiments.  This is because factors that reduce the efficiency of the real robots digging (due primarily to their design) such as side spillage of soil while pushing or frequently getting stuck while pushing results in less soil being transferred to the intended destination.  Furthermore, other in-situ factors may further reduce excavation efficiency of each robot on the lunar surface.      As a result, each robot requires more time to reach a steady state fitness.   One solution would be to supply the optimal number of robots for the given work area as described earlier.  This would further parallelize the task and enable task completion in shorter time despite any limitations in design.  Given higher power rovers, the robots could conceivably achieved the required goal map in shorter time.
Nevertheless our results show that a simple simulation environment, with low-cost proof of concept vision hardware is sufficient to demonstrate the critical behaviors.

Table~\ref{tb:excavate_metric} shows nearly 70\% of the area is excavated towards the goal depth or material dumped to correct regions according to the goal map, while nearly 24\% had material accumulated incorrectly (either regolith dumped at locations where it is suppose to be dug out or regolith dug out where it is to dumped).  These results were achieved with low cost vision hardware, principally webcams and single-pixel laser rangers.  The results suggest that the robot controllers correctly interpret and follow the goal map on most occasions.  However a series of real world conditions imparts some error to the results particularly in dumping regolith off away from the dumping areas due to drift errors.  This provides 1-3 cm error in position of the dumping areas.  This accounts for a significant portion of the error.  These drift errors while lowering fitness metrics are actually not a major concern for field applications.  These field experiments would suggest we introduce some margin distance between critical boundaries that can be easily included into the blueprints. Other factors that contributed to errors in the experiments, include the overhead system at times loosing track of robot position.

\begin{table*}[t!]
\caption{Field Experiment Metrics } \label{tb:excavate_metric}
 \vspace{2pt}
\centering
\footnotesize{
\begin{tabular}{c l }
\hline
{\bf Area}           & {\bf System Response}   \\
\hline\hline
72\%               & Correctly Excavated towards Goal Depth  \\
24\%               & Above Goal Depth  \\
4\%                & Indeterminate \\
\hline
\end{tabular}}
\end{table*}

\begin{figure} [H]
    \centering
    \includegraphics[width=4.0in]{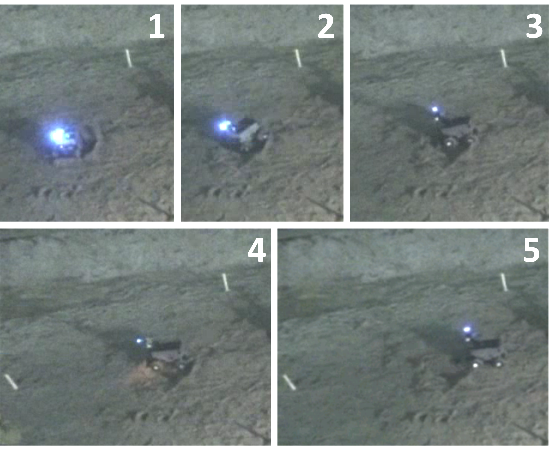}
  %   \vspace{-12pt}
\caption{Video snapshots of the stuck avoidance behavior, including a rocking behavior followed by a `backout' behavior triggered after the robot gets stuck trying to push the blade forward.} \label{fig:stuck_impressive}
% \vspace{-5pt}
\end{figure}

\begin{figure} [H]
    \centering
    \includegraphics[width=6.2in]{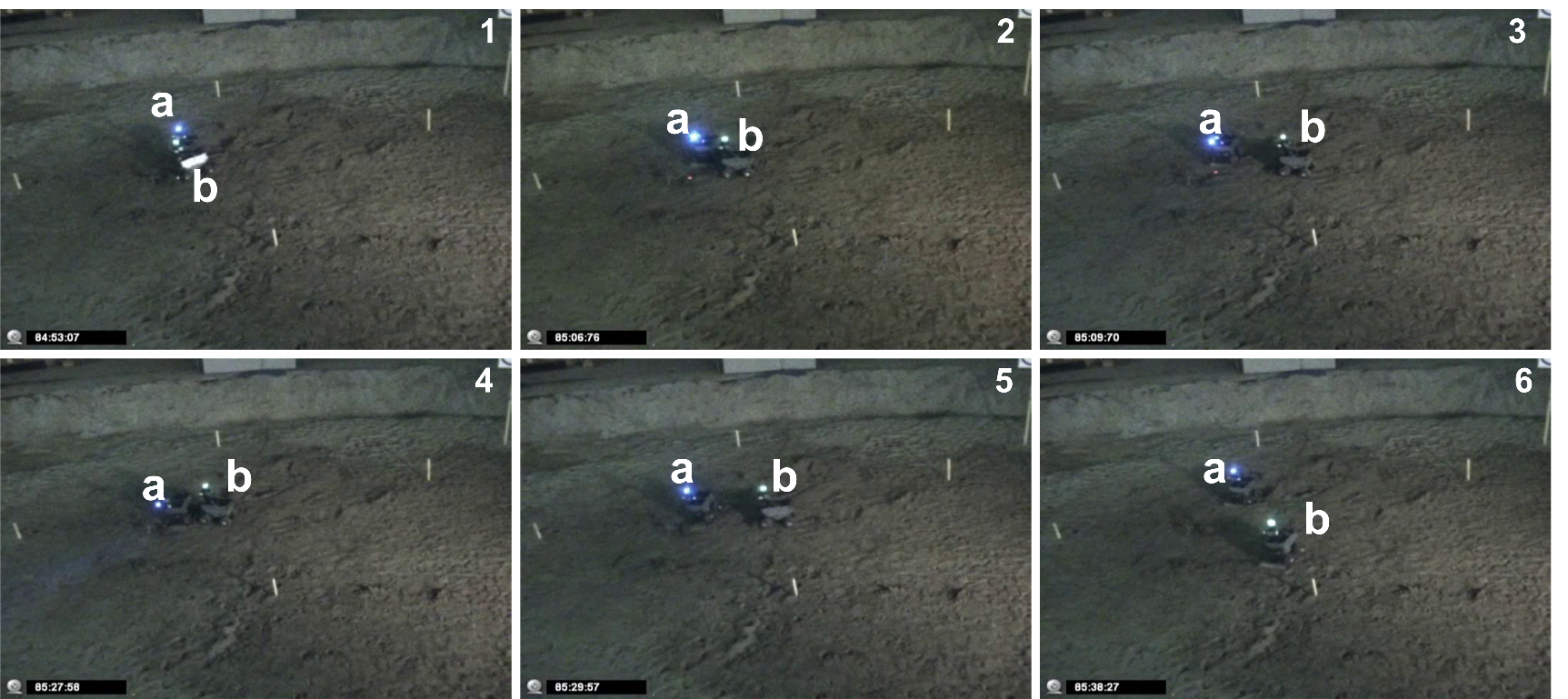}
   %  \vspace{-12pt}
\caption{ANT controllers evolve obstacle avoidance/negotation behaviors.} \label{fig:obstacle_avoidance}
%\vspace{-5pt}
\end{figure}

The field experiments show several creative behaviors discovered by the controllers in solving the task.  This includes correctly reading templates and cues in the environment such as the excavation blueprint, rocking and obstacle avoidance behaviors.  The controllers evolve rocking behavior to get unstuck, a method of going back and forth to dislodge itself from sand traps and deep `pot holes.'  In this scenario, the controller learns to stop repeated excavation moves and `back out' when it is unfruitful to dig or when the robot is stuck moving forward (Figure~\ref{fig:stuck_impressive}).  This is accomplished using the memory variable to represent the `stuck state' and prevent repeated attempts at digging.  In addition, the controllers evolve obstacle avoidance/negotiation behavior.  Figure~\ref{fig:obstacle_avoidance} shows video snapshots of this behavior.  Two robots are attempting to excavate and push dirt towards the dumping area. In frame 1, the robot in front (b) makes a left turn but is in the way of robot (a). Robot (b) backs out and scans in front (frame 2).  It detects robot (a) and backs up further (frame 3).  In frame 4, robot (b) attempts to move forward assuming the path is clear and stops having detected robot (a). In frame 5, robot (b) backs up again, makes a right turn and moves forward facing the dumping area as originally intended.  The obstacle avoidance behaviours from high-fidelity simulations and hardware experiments show that the robots avoid dumping soil on or next to each other preventing the robots from getting stuck.  This suggests the controllers use proximity sensing to avoid such actions.

These field experiments demonstrate proof-of-concept feasibility of applying ANT on multirobot excavation for lunar applications using low-cost vision hardware consisting of webcams and laser range finder.  However significant work remains, particularly in demonstrating large-scale excavation. Work is underway in developing a team of large excavation robots for a field demonstration. Another solution is to use other excavation implements such as a bucket wheel or front loaders and is being actively explored using this approach.

\section{Conclusion}

The ``Artificial Neural
Tissue'' (ANT)
framework, a neural network approach to robotic control has been  applied to multirobot excavation.  Using this approach we show the feasibility of using multirobot excavation
for site preparations tasks.  The approach shows improved performance and scalability than conventional neural networks and hand coded solutions.   This facilitates finding creative behaviors that are not specified or encouraged by an experimenter.  These creative behaviors verified in hardware include correctly interpreting blueprints, performing layered digging, obstacle avoidance and rocking behaviors to avoid getting stuck.  This approach is shown to produce controllers that have improved scalability compared to conventional neural networks and hand-coded solutions.Furthermore, ANT can simultaneously evolve the desired controller and select for optimal number of robots for the task. This approach is shown as a possible solution to the problem of antagonism in decentralized multirobot control.

This approach shows that a machine learning method such as ANT is a machine driven process  to develop creative yet effective deterministic methods for multirobot excavation. ANT evolves  efficient excavation methods such as slot dozing that would otherwise require experts.   This is important in space operations, where the space system operational  engineer may not be a excavation expert.  It is fair to say ANT is unlikely to outperform a trained excavation vehicle operator in the field.  But the ability for the an autonomous algorithm to approach that capability is significant for off world excavation, where it is prohibitively expensive to send a full team of astronauts to perform excavation.  The proposed approach would rely on combination of teleoperation and autonomous control to perform excavation.  The proposed approach would automate mundane excavation and regolith moving tasks.   There would however be an important need for oversight, where a teleoperator monitors the excavation process and intervenes when there is unexpected failures in the field.  Thanks to the use of the proposed  autonomous control approach to perform most of the time consuming and mundane tasks, this can reduce the number of teleoperators and also reduce fatigue due to communication latency.  However significant work is required in infusing teleoperation based control with autonomous control.

Future work is planned for full field excavation experiments using multiple robots.  This work will combine both teleoperation and autonomous control.  The proposed approach requires effective methods of surveying and localization to position the excavator and achieve correct dimensions of features in the work site.  Our experiments demonstrate use of overhead cameras to perform localization.  Use of overhead cameras is a viable approach on the moon.  In addition, radio beacons may also be used for localization.

Overall, the proposed approach shows a credible pathway towards lunar excavation, ISRU and base construction.  Further work is needed in fusing teloperation with proposed autonomous controls approach.  Challenges remain in scaling up the rovers and localization system to an appropriate size for a representative field demonstration.

\subsection{Acknowledgements}

This research has been supported by Natural Science and Engineering Research Council of Canada, NORCAT and EVC. The authors would gratefully acknowledge the contribution of Dale Boucher of NORCAT and Jim Richards of EVC.  The authors would like to thank the reviewers for helping better identify the strengths and weaknesses in the proposed approach.

%\begin{acknowledgements}
%\end{acknowledgements}


\bibliographystyle{apalike}
\bibliography{robotica_submission}

\section{Appendix}

\subsection{Hand-Coded Controller}

The handed-coded controller consists of 11 if-then statements executed at each timestep on each robot.  The descriptions here mention which output behaviors are selected according to those given in Table 2.
 \vspace{12pt}

\noindent 1) Checks if the robot is stuck and selects behaviors 1, 4, 7, 10, 12.

\noindent 2) Otherwise if the robot faces an obstacle, it selects behavior 4.  Otherwise, the remaining if statements are executed as follows.

\noindent 3) If region $Z_2$ or $Z_3$ are in the ``dumping zone'' and the robot is pushing soil,  $L_1 > 0$ and selects behavior 2, 3, 4.

\noindent 4) If region $Z_2$ or $Z_3$ are in the ``don't care'' and the robot is pushing soil,  $L_1 > 0$ and selects behavior 3, 4, 12.

\noindent 5) If region $Z_2$ or $Z_3$ are in the ``don't care'' then selects behaviors 2, 4.

\noindent 6) If region $Z_2$ or $Z_3$ are in the ``don't care'' and  $L_1 = 0$ then select behavior 3.

\noindent 7) If region $Z_2$ or $Z_3$ are at goal depth, then behaviors 2, 4 are selected.

\noindent 8) If region $Z_2$ or $Z_3$ are below goal depth, then behaviors 2, 7, 11 are selected.

\noindent 9) If region $Z_2$ or $Z_3$ are above goal depth, then behaviors 2 is selected.

\noindent 10) If region $Z_2$ or $Z_3$ are above goal depth and memory bit is 0, then behaviors 8, 11 are selected.

\noindent 11) If region $Z_2$ or $Z_3$ are above goal depth and memory bit is 1, then behaviors 2, 9 are selected.

\vspace{12pt}

All selected behaviors are executed sequentially according to Table 2.  The selections are then reset and control program executed on each robot for the next timestep.

\subsection{Sensor-Behavior Detectors}
Table 3 lists the sensor behavior detectors used analyze the evolution of behaviors in the ANT controllers.

\begin{table*}[h!]
\caption{Sensor-Behavior Detectors} \label{tbl:detect}
 \vspace{2pt}
\centering
\footnotesize{
\begin{tabular}{c l l}
\hline
{\bf Detector}           & {\bf Name}      & {\bf Logic}                                                \\
\hline\hline
                &                  &                                    \\
1               & Level            & If $B_1$ = Level and $L_1 > 0$ and Move Forward = 1                       \\
                &                  &                                    \\
2               & Collision Avoidance   & If $S_1 = 1$ and Move Forward = 0 and Random Turn = 0  and                       \\
                &                       &(Turn Right $\neq$ Turn Left or Move Backward = 1)                                \\
                &                  &                                    \\
3               & Stuck Avoidance  & If $U_1 = 1$ and Move Forward = 0 and (Move Backward = 1 or Random Turn = 1 or                              \\
                &                  & Turn Left $\neq$ Turn Right)                              \\
                &                  &                                    \\
4               & Cut Dig          & If $B_1$= Below and Move Forward = 1                              \\
                &                  &                                                            \\
5               & Correct Dump     & If $Z_2$ = Dump and $Z_3$ = Dump and $L_1 > 0$ and Random Turn = 0 and                               \\
                &                  & Move Forward = 1 and  Turn Left = Turn Right                               \\
                &                  & or                               \\
                &                  & If $Z_2$ = Dump and $Z_3$ = Dump and $L_1 > 0$ and Random Turn = 1 and                               \\
                &                  & Random Turn $\longrightarrow$ Turn Left and Turn Right = 1 and and Turn Left = 0 and                                 \\
                &                  & Move Forward = 1                               \\
                &                  & or                               \\
                &                  & If $Z_2$ = Dump and $Z_3$ = Dump and $L_1 > 0$ and Random Turn = 1 and                               \\
                &                  & Random Turn $\longrightarrow$ Turn Right and Turn Right = 0 and and Turn Left = 1 and                                \\
                &                  & Move Forward = 1                               \\
                &                  &                                    \\

\hline
\end{tabular}}
\end{table*}

\end{document}